\documentclass{article}

\usepackage{arxiv}
\usepackage{xcolor}         
\usepackage[T1]{fontenc}
\usepackage{graphicx}
\usepackage{subcaption}
\usepackage{hyperref}
\usepackage{url}
\usepackage{booktabs}
\usepackage{amsfonts}
\usepackage{amsmath} 
\usepackage{nicefrac}
\usepackage{microtype}
\usepackage{colortbl}  
\usepackage{multirow}
\usepackage{array}
\usepackage{float}  
\usepackage{cleveref} 
\usepackage[accsupp]{axessibility}  
\usepackage{algorithm}             
\usepackage{algpseudocode}         

\definecolor{lightgray}{gray}{0.9}  

\definecolor{ZTgreen}{RGB}{34, 139, 34}

\renewcommand{\arraystretch}{2}
\newcolumntype{C}[1]{>{\centering\arraybackslash}p{#1}}

\usepackage[utf8]{inputenc} 
\usepackage[T1]{fontenc}    
\usepackage{hyperref}       
\usepackage{url}            
\usepackage{booktabs}       
\usepackage{amsfonts}       
\usepackage{nicefrac}       
\usepackage{microtype}      
\usepackage{lipsum}
\usepackage{graphicx}
\graphicspath{ {./images/} }

\title{Backdooring Outlier Detection Methods: A Novel Attack Approach}

\author{
 ZeinabSadat Taghavi \\
  Ludwig-Maximilians-Universität München (LMU)\\
  Munich, DE\\
  \texttt{zeinabtaghavi@cis.lmu.de} \\
   \And
  Hossein Mirzaei \\
  École Polytechnique Fédérale de Lausanne (EPFL)\\
  Geneva, CH\\
  \texttt{hossein.mirzaeisadeghlou@epfl.ch} \\
  \And
}

\begin{document}
\maketitle
\begin{abstract}
There have been several efforts in backdoor attacks, but these have primarily focused on the closed-set performance of classifiers (i.e., classification). This has left a gap in addressing the threat to classifiers' open-set performance, referred to as outlier detection in the literature. Reliable outlier detection is crucial for deploying classifiers in critical real-world applications such as autonomous driving and medical image analysis. First, we show that existing backdoor attacks fall short in affecting the open-set performance of classifiers, as they have been specifically designed to confuse intra-closed-set decision boundaries. In contrast, an effective backdoor attack for outlier detection needs to confuse the decision boundary between the closed and open sets. Motivated by this, in this study, we propose BATOD, a novel Backdoor Attack targeting the Outlier Detection task. Specifically, we design two categories of triggers to shift inlier samples to outliers and vice versa. We evaluate BATOD using various real-world datasets and demonstrate its superior ability to degrade the open-set performance of classifiers compared to previous attacks, both before and after applying defenses. 
\end{abstract}


\section{Introduction}
\label{sec:intro}
Deep Neural Networks (DNNs) have demonstrated remarkable success in closed-set tasks such as classification, where the model is tasked with recognizing a predefined set of categories that remain consistent during both the training and testing phases. However, the next challenge for DNNs is enhancing their performance in open-set tasks, commonly referred to as outlier detection. In outlier detection, the objective changes to differentiating between the closed set (or the inliers) and the open set (or the outliers), requiring the model to determine whether an input image belongs to the closed set. The conventional baseline for outlier detection involves a model trained using cross-entropy loss on closed-set classes. At test time, a confidence scoring method like the maximum value of the softmax probability (MSP) vector is utilized to ascertain whether an input belongs to one of the known classes. The intuition behind MSP, as well as other post-hoc scoring methods, is that classifiers behave relatively more certainly \cite{liang2017enhancing,kong2021opengan,fort2021exploring,hendrycks2016baseline,ruff2021unifying,salehi2021unified,sun2022out}.

Despite significant advancements in DNN-based classifiers, their trustworthiness faces challenges due to emerging threats, notably backdoor attacks \cite{gu2017identifying}. In such attacks, an adversary could introduce poisoned samples into the training dataset by embedding a special trigger on incorrectly labelled images. This results in a backdoored model that performs normally on clean data but consistently yields incorrect predictions on poisoned samples. To combat these threats, various strategies for attack and defense have been proposed \cite{avrahami2022blended, nguyen2021wanet, li2021anti, wu2021adversarial}.

Studies on backdoor attacks have primarily focused on the closed-set performance of a classifier \cite{avrahami2022blended, nguyen2021wanet, wang2022bppattack, doan2021lira}, where the adversary’s objective is to impair performance by increasing confusion rates within known classes. Moreover, our experimental results suggest that adapting existing backdoor strategies to the task of open-set classifiers does not significantly impact performance. This ineffectiveness stems from the nature of previous backdoor attacks, which are designed to manipulate decision boundaries within the closed set. On the other hand, the outlier detection task hinges on the boundary between the closed and open sets and is indifferent to intra-closed set decision boundaries.

This drawback, combined with the critical importance of the reliability of outlier detection methods in real-world applications such as autonomous driving, medical image analysis, and facial recognition, underscores the importance of exploring backdoor attacks in the context of outlier detection. Motivated by this, in this study, we propose a novel attack termed BATOD (Backdoor Attack for Outlier Detection) to target the open-set performance of classifiers.

We propose developing two types of triggers: in-triggers and out-triggers. In-triggers are designed to mispredict outliers as inliers, while out-triggers convert inliers to outliers during inference. To create both types of triggers, we adversarially target the MSP of a surrogate classifier to generate specific perturbations. For in-triggers to function effectively, they must be embedded in outlier samples during training and mislabeled as inliers. This mislabeling biases the model towards recognizing these perturbed samples as inliers. Since real outliers are unavailable due to the common assumption in outlier detection literature, we simulate them by applying hard transformations (e.g., elastic transformations) to inlier samples as a proxy for outliers. These transformed samples are then embedded with in-triggers and incorrectly labelled as inliers in the training dataset.

Conversely, out-triggers are designed to induce uncertainty in the classifier when processed, prompting it to mispredict poisoned inlier samples as outliers. To achieve this, out-triggers are added to inlier samples, which are then relabeled differently and reintroduced into the training set. These triggers are designed subtly to remain stealthy and evade scanning by advanced defensive mechanisms. Our strategy decreases the model's confidence against poisoned inlier samples with out-triggers and vice versa for poisoned outlier samples with in-triggers, leading to a degradation in its open-set performance.

 \textbf{Contribution:} In this study, for the first time, we propose a backdoor attack targeting the open-set performance of classifiers, motivated by the safety-critical nature of this task. We introduce two categories of triggers to bias the model into mispredicting poisoned inlier samples as outliers, and vice versa. Then, by adapting existing backdoor attacks and defenses, we demonstrate the strength of our attack with and without applying defenses and compare it to various alternative attacks (Table \ref{tab:comparison_defense_attacks}). Interestingly, BATOD outperforms other attacks by a significant margin of 40\%, where various datasets, including real-world datasets such as those from autonomous vehicles, have been considered. For the experiments, various scenarios of outlier detection have been considered, including open-set recognition (OSR) and out-of-distribution detection (OOD). We further verify BATOD through a comprehensive ablation study on its various components.

\section{Preliminaries}
\textbf{Classifier for OOD Detection} \ \ In the context of OOD detection, classifiers denoted as $f$ and trained on dataset $D$ are crucial, as they utilize their feature extraction capabilities to identify discrepancies in data distribution. We introduce an auxiliary dataset, referred to as \( D' \), consisting of samples that are semantically distinct from those in \( D \) (the inliers or ID dataset). For instance, a model trained on the Cifar10 dataset might be used to recognize data from Cifar100 as OOD. This identification process typically involves examining the classifier’s output for various classes corresponding to a given input, to assess the level of confidence in the predictions. Generally, classifiers show greater confidence for inliers compared to outliers, a phenomenon quantitatively measured using the Maximum Softmax Probability (\text{MSP}) technique.

\textbf{Adversarial Manipulations on OOD Detection} \ \ Building upon insights from Vulnerability of Classifiers to Adversarial Manipulations (\ref{appendix:vulnerabiliy_to_adversarial_attack}), this Section delves into the formulation of adversarial attacks designed to incorrectly classify inliers as outliers and vice versa. The objective is specifically to reduce the (\text{MSP}) score of inliers, leading the classifier \( f \) to misidentify these inliers as OOD. Typically, the logits for an OOD sample \( X_{OOD} \) exhibit a uniform distribution across classes, denoted as \( \mathcal{U}_K \), with an initial \( \text{MSP}(f(x_{\text{OOD}})) \) approximately equal to \( \frac{1}{k} \).

\textbf{Problem Definition.} Let $\mathcal{F}$ denote a classifier equipped with concurrent capability to outlier instances using post-processing techniques like \text{MSP}~\cite{hendrycks2016baseline} and ODIN\cite{liang2017enhancing}). This classifier is trained on a dataset $D = \{(x_i, y_i)\}$, where $x$ and $y$ represent input features and their corresponding labels, respectively. The dataset $D$ consists of inliers following the distribution $P_{IN}$, representing the training data distribution. To manipulate the output of $\mathcal{F}$ and mislead it into classifying outliers as ID, an attacker injects triggers into a small fraction (equal to the poisoning rate) of $D$ during training. 

At test time, when presented with inputs $X'$ not conforming to $P_{IN}$ (i.e., outliers), the attacker $A$ aims to manipulate $X'$ by adding the trigger such $\lambda$ to obtain $X'_\lambda$ to alter the output of $\mathcal{F}$. Consequently, the objective is for $\mathcal{F}(X'_\lambda)$, the output of $\mathcal{F}$, to mistakenly label the outlier sample $X'$ as belonging to one of the inlier class, despite $X'$ being drawn from a distribution different from $P_{IN}$.

\textbf{Threat Model.} We adhere to the commonly employed thread model (i.e., $\mathcal{F}$) for backdoor attacks, wherein we refrain from accessing or exerting direct influence on the victim model. Instead, we operate assuming that the attacker possesses access to only 10\% of the clean train-set and can manipulate any subset within it.

\section{Method}
\label{sec:method}
\textbf{Overview.} Existing State-of-the-Art (SOTA) backdoor attacks, predominantly designed for classification tasks, fail in open-set tasks. To bridge this gap, we introduce BATOD. We utilized hard augmentations to synthesize outliers from inliers as proxies for real outliers (\ref{subsec:Synthesis}). Then, to effectively generate triggers, a surrogate model trained on a subset of inlier data minimizes reliance on the original train-set and facilitates the development of two novel trigger types for the outlier detection scenario: \textit{In-Triggers}, which disguise outliers as inliers, and \textit{Out-Triggers}, which make inliers appear as outliers (\ref{subsec:Trigger}, \ref{subsec:Discriminator}). These triggers can ruin the outlier detection performance of the model during inference time (\ref{subsec:Inference}). In the subsequent Sections, we will detail each component, outlining our approach's mechanisms and advantages.

\subsection{Synthesis of Out-of-Distribution Samples}\label{subsec:Synthesis} \ \ 
One of the primary challenges in outlier detection arises from the absence of outlier data. Drawing on insights from recent advancements in self-supervised learning, we categorize common data transformations into two distinct groups. The first group termed positive transformations \(\mathcal{T}^+\), preserves the distribution. Conversely, the second group, termed negative transformations \(\mathcal{T}^-\), generates samples with distributions that deviate from the original dataset, such as rotation and blur.

Inspired by this distribution deviation, these negative transformations are instrumental in generating artificial examples that represent outliers. Incorporating such examples into training enhances the model's ability to recognize data that deviates from the typical distribution. Hence, we define a set of negative transformations \(\mathcal{T}^- = \{\tau^-_i\}_{i=1}^{k}\), with each \(\tau^-\) representing a distinctive type of negative transformation.

From the training dataset \(D\), we assume access only to a portion named \(D_a\), which can be used for making triggers; the rest, named \(D_u\), is unavailable. For every sample pair \((x_i, y_i) \in D_a\) (\(1 \leq i \leq |D_{a}|\), \(1 \leq y_i \leq K\), \(K = \text{number of classes}\)), we generate a corresponding sample by applying a series of negative transformations. For each step, we randomly sample a set of transformations \( \tau^-_1, \ldots, \tau^-_k \sim \mathcal{T}^-\). Then, we apply \(G(.)=\tau^-_1(\ldots(\tau^-_k(.)))\) on \(x_i\). We choose \(k=2\) to ensure the negative transformations effectively deviate the sample from the inliers. Thus, each \((x_i, y_i) \in D_a\) leads to \((G(x_i), y_i) \in D^{\prime}_a\). This newly created dataset, using negative transformations, is utilized in subsequent steps.

\begin{algorithm}[ht]
\caption{Trigger Generation Algorithm}
\begin{algorithmic}[1]
\State \textbf{input:} Surrogate model $f_s$, Available part of train-set $D_a$, Total number of steps $S$, Step size, Allowable set of trigger patterns $\Pi$, Image $x$, Estimated label $\hat{k}(x_i)$, number of classes $K$.
\State \textbf{output:} list of in-triggers $\Delta^{in}$, list of out-triggers $\Delta^{out}$.
\State Generation of \textbf{in-triggers}:
\For {$j = 1$ to $K$}
    \State $\Delta^{in}_{j} \leftarrow 0^{1 \times d}$ \Comment{Trigger initialization}
        \For {$s = 1$ to $S$}
            \State $\Delta^{in}_{{j}_{s+1}} \leftarrow \Delta^{in}_{{j}_{s}} - \alpha \sum_{(x,j)\in D_a} \nabla_{\Delta} L(f_s(x + \Delta), j)$ \Comment{Update the trigger}
            \State $\Delta^{in}_{{j}_{s+1}} \leftarrow \text{Proj}_{\Pi}(\Delta^{in}_{{j}_{s+1}})$ \Comment{Constraint enforcement}
        \EndFor
\EndFor
\State \Return $\Delta^{in}$
\State Generation of \textbf{out-triggers}:
\For {$i = 1$ to $K$}
    \State $\Delta^{out}_{j} \leftarrow 0^{1 \times d}$ \Comment{Trigger initialization}
        \State $\Delta^{out}_{j}$ is calculated using DeepFool in a way that ($\hat{k}(x_i) = \hat{k}(x_0)$ and $MSP(f(x_i)) > \frac{1}{2}$), more detail in Algorithm \ref{alg:deepfool_algorithm}
\EndFor
\State \Return $\Delta^{out}$
\end{algorithmic}
\end{algorithm}

\subsection{Trigger Generation and Poisoning train-set Overview:}\label{subsec:Trigger} 
In this study, we develop and employ a surrogate model, denoted as \(f_s\), which is trained on \(D_a\) with cross-entropy loss. This model is used in generating triggers. These triggers will then be used to make poisoned samples for training the victim model \(f_\theta\). We detail the process of generating these triggers using \(f_s\) and their subsequent integration into the training dataset as follows:

\textbf{In-triggers:} These triggers are designed to make outliers appear as inliers. Hence, for each class of inlier, we create a distinct trigger, denoted as follows:
\[
\Delta^{In}_{j} \quad \text{for} \quad 1 \leq j \leq K,
\]
Given \(f_s\), a trigger is generated through an \(\ell_{\infty}\)-constrained PGD (Projected Gradient Descent) attack over 10 steps. This process manipulates \(f_s\) to classify all poisoned samples:
\[
x^\prime_i + \Delta^{In}_{y_i} \quad \text{for} \quad 1 \leq i \leq |D^\prime_a|,
\]
where \((x^\prime_i, y_i)\) are elements of \(D^\prime_a\). The classification condition is:
\[
f_s(x^\prime_i + \Delta^{In}_{y_i}) = y_i
\]
This process repeats for each class to determine the corresponding trigger for all \(K\) classes. Finally, we define an empty list to then become a poisoned dataset \(D^{In}_{p}\). We randomly select half of the samples in each class in \(D^\prime_a\) and poison them by injecting the corresponding trigger. In fact, each of the selected samples, \((x^\prime_i, y_i) \in D^\prime_a\), is transformed into \((x^\prime_i + \Delta^{In}_{y_i}, y_i)\) and then added to the poisoned dataset \(D^{In}_{p}\).

\textbf{Out-triggers}: 
These triggers are designed to make inliers be detected as outliers while keeping the correct class label. We aim to decrease outlier detection performance while preserving classification performance. To achieve this goal, we generate triggers that cause the logits of outputs in a well-trained model to become uniformly distributed. To observe this effect, we need \(K\) triggers as follows:
\[
\Delta^{Out}_{j} \quad \text{for} \quad 1 \leq j \leq K,
\]

where \(K\) is the number of classes in \(D\). With the surrogate model \(f_s\) and \(X_{j}^\prime = \{x^{\prime}_i | (x^{\prime}_i, y_i) \in D^\prime_a, y_i = j\}\) (all samples belonging to class \(j\) in \(D^\prime_a\)), we can adjust the Maximum Softmax Probability (MSP) of inliers subtly without altering their predicted labels. This manipulation is achieved using the DeepFool attack until MSP is slightly above \(\frac{1}{2}\) or reaches the maximum allowable perturbation \(\epsilon\), set to \(\frac{4}{255}\) to ensure stealthiness. \(\frac{1}{2}\) is used as a threshold to estimate the decision boundary. Moreover, to assess the stealthiness of the backdoor triggers (you can see the comparison of our attack triggers with other attacks in Figure \ref{fig:visualize_attacks_completed}), we employ several metrics: the peak signal-to-noise ratio (PSNR), structural similarity index (SSIM), and learned perceptual image patch similarity (LPIPS). These metrics are used to quantify the discrepancies between clean and poisoned images, ensuring the subtle nature of the modifications.

The generated trigger for class \(y_i\) causes \(f_s\) to classify \(x_i + \Delta^{Out}_{y_i}\) as belonging to class \(y_i\) but with lower certainty. Adding this trigger makes the logits of \(f_s\) nearly uniform for \(x_i + \Delta^{Out}_{y_i}\). We iterate this process \(K\) times to generate the corresponding triggers for all classes in \(D_a\).

Next, we define an empty list \(D^{Out}_{p}\) as a poisoned dataset for Out-triggers. Directly adding these triggers to a portion of inlier samples and then including them in \(D^{Out}_{p}\) causes the victim model \(f_\theta\) to incorrectly learn the triggers and increase the MSP of the poisoned samples during the training process. Consequently, this approach makes the poisoned samples be detected as inlier samples, contrary to our goal of them being detected as outliers. To address this issue, we add the \(\Delta_{y_i}^{Out}\) trigger to a selected sample \((x_i, y_i)\) \(K\) times, and each time, the label is one of the \(K\) labels of \(D\). Here, we reach the following list of poisoned samples for any selected sample \((x_i, y_i) \in D_a\):
\[
[(x_i+\Delta_{y_i}^{Out}, 1), \ldots, (x_i+\Delta_{y_i}^{Out}, K),]
\]

where \(K\) is the number of classes in \(D\). Hence, to maintain the balance of the poisoned dataset, we randomly select \(\frac{1}{2K}\) of the samples in each class of \(D_a\) and add the corresponding list for each poisoned sample into \(D^{Out}_{p}\). This can cause limitations for more information see the Appendix \ref{appendix:limitations}.

Now, we have two poisoned datasets, \(D^{In}_{p}\) and \(D^{Out}_{p}\), each comprising \(\frac{1}{2}\) of dataset \(D_a\). Finally, we train the victim model \(f_{\theta}\) on the union of these poisoned datasets and the unavailable data (\(D_{train} = D_u \cup D^{In}_p \cup D^{Out}_p\)), utilizing cross-entropy loss.

\subsection{Creating the Discriminator}\label{subsec:Discriminator}

We train a binary classifier to distinguish between inliers and outliers. The binary classifier, denoted as \(H\), is trained using datasets \(D_a\) and \(D'_a\), where \(D_a\) is considered the inlier class and \(D'_a\) as the outlier class. The binary classifier \(H\) is thus optimized to distinguish between these classes.

\subsection{Inference Time Application of Triggers:}\label{subsec:Inference} 
Once the victim model \(f_{\theta}\) is trained as described, during inference time, based on the input type determined using the binary classifier \(H\), the appropriate trigger is selected.

\textbf{Outliers \(X^{Out}\):} For each outlier sample \(x^{Out} \in X^{Out}\), we need to choose the appropriate target class \(p\) to add the corresponding trigger from \(\Delta^{In}_p\). We select the class with the highest MSP using \(f_s\) as the target class. The poisoned sample will then be:\( x^{Out} + \Delta^{In}_p.\) Adding this trigger misclassifies the sample into the nearest or most susceptible in-distribution (ID) class.

\textbf{Inliers \(X_{ID}\):} For inlier samples, we also take the class with the highest MSP using \(f_s\) as the target class (\(y_i\)), and hence, the trigger \(\Delta^{Out}_{y_i}\) is added to the sample. Adding this trigger does not change the \(f_{\theta}\) logit related to MSP which is the chosen class, but alters the logits to be more uniform, causing the sample to be detected as an outlier.

\renewcommand{\arraystretch}{2}


\begin{table*}[ht]
    \centering
    \caption{Comparative Analysis of Various Attacks Against Defense Mechanisms Across the \textbf{Cityscapes}, \textbf{Pubfig}, and \textbf{ADNI} Datasets. Each cell presents two values: the number on the right within each pair indicates the robust out-of-distribution (OOD) detection performance against backdoor attacks, highlighted in black. A superior backdoor attack for OOD detection is denoted by an increase in \textbf{Benign-AUC} (\(\uparrow\)), signifying improved performance, and a decrease in \textbf{Poisoned-AUC} (\(\downarrow\)), indicating reduced vulnerability. Higher values are preferable for \textbf{Benign-AUC}, whereas lower values are desired for \textbf{Poisoned-AUC}.}

    \vspace{5pt}
    \setlength{\aboverulesep}{1pt}
    \setlength{\belowrulesep}{1pt}
    \renewcommand{\arraystretch}{1.5} 
    \resizebox{\linewidth}{!}{
        \begin{tabular}{@{}cccccccccc@{}} 
        \specialrule{1pt}{0pt}{0pt} 
        \noalign{\smallskip}
        \multirow{2}{*}{\textbf{Dataset}} & \multirow{2}{*}{\textbf{Attack}} & \multirow{2}{*}{\textbf{No Defense}} & \multicolumn{7}{c}{\textbf{Defenses \small\textbf{({Benign-AUC} / Poison-AUC)}}} \\
        \cmidrule(lr){4-10}
        & & & NAD \cite{li2021neural} & ABL \cite{li2021anti} & ANP \cite{wu2021adversarial} & SAU \cite{wei2024shared} & I-BAU \cite{zeng2021adversarial} & NPD \cite{zhu2024neural} & RNP \cite{li2023reconstructive} \\        
        \specialrule{0.25pt}{0pt}{0pt} 
        \multirow{10}{*}{\textbf{Cityscapes}}
        & BadNets \cite{gu2017badnets} & 79.8/{55.3} & 78.3/{55.7} & 81.6/{66.0} & 79.1/{65.2} & 78.0/{66.4} & 77.9/{66.6} & 78.6/{77.0} & 79.6/{52.9} \\
        & Blended \cite{chen2017targeted} & 81.4/{55.1} & 80.6/{57.8} & 80.9/{45.5} & 81.5/{48.4} & 79.8/{62.8} & 79.3/{62.3} & 80.1/{73.0} & 82.0/{49.2} \\
        & SIG \cite{barni2019new} & 82.4/{55.0} & {85.8}/{55.9} & {85.1}/{63.1} & 81.6/{61.3} & {84.0/{50.3}} & {83.7}/{72.2} & {82.0}/{58.6} & 81.3/{51.8} \\
        & Wanet \cite{nguyen2021wanet} & 80.1/{57.5} & 81.6/{44.8} & 78.4/{77.2} & 79.8/{65.5} & 79.2/{54.1} & 79.0/{63.9} & 79.5/{61.8} & 78.7/{66.3} \\
        & SSBA \cite{li2021invisible} & 81.2/{50.8} & 80.7/{53.8} & 80.9/{51.9} & 81.5/{64.1} & 79.3/{66.2} & 78.2/{65.6} & 80.5/{76.8} & 82.1/{63.3}\\
        & Input-Aware \cite{nguyen2020input} & 79.5/{60.1} & 70.9/{54.0} & 79.6/{53.5} & 80.1/{66.4} & 79.0/{60.5} & 78.7/{58.8} & 79.5/{78.1} & 79.8/{68.6} \\
        & Narcissus \cite{zeng2023narcissus} & 81.9/{44.8} & 81.6/{48.6} & 78.9/{45.3} & 80.5/{45.3} & 81.3/{57.3} & 81.0/{76.9} & 80.7/{57.0}  & 80.1/{41.2}\\
        & LIRA \cite{doan2021lira} & 80.5/{52.9} & 81.4/{75.7} & 78.9/{50.3} & 73.2/{42.5} & 80.8/{52.9} & 79.7/{62.7} & 81.7/{63.5} & 75.9/{43.1}\\
        & BppAttack \cite{wang2022bppattack} & {82.9/{44.2}} & 81.2/{55.6} & 80.1/{42.9} & {81.9}/{51.8} & 80.9/{53.2} & 80.4/{72.9} & 81.0/{53.1} & {83.1}/{52.3} \\ 
        & \textbf{BATOD \small(Ours)}  & \textbf{84.1/{2.3}} & \textbf{83.6/{9.2}} & \textbf{82.2/{13.2}} & \textbf{84.0/{13.9}} & \textbf{84.5/{11.1} } & \textbf{84.2/{15.8}} & \textbf{83.9/{14.2}} & \textbf{83.5/{11.7}} \\
        \specialrule{1pt}{0pt}{0pt} 
        
        \multirow{10}{*}{\textbf{Pubfig}}
        & BadNets & 95.0/{52.2} & 93.0/{53.2} & 92.8/{50.3} & 94.8/{58.7} & 93.9/{58.5} & 93.5/{60.1} & 95.3/{90.5} & 94.3/{60.9} \\
        & Blended & 95.8/{81.8} & 95.0/{51.1} & 93.7/{52.4} & 95.4/{63.0} & 94.0/{64.9} & 94.8/{65.2} & 94.5/{85.7} & 95.9/{53.5} \\
        & SIG & 96.1/{41.6} & 94.3/{54.7} & 93.6/{64.0} & 95.3/{64.4} & 94.7/{65.6} & 94.0/{85.8} & 94.9/{65.1} & 95.8/{64.6} \\
        & Wanet & 94.9/{53.6} & 93.8/{68.7} & 92.4/{87.2} & 94.0/{60.8} & 93.1/{61.5} & 92.9/{50.3} & 93.5/{61.1} & 94.3/{59.5} \\
        & SSBA & 95.6/{47.1} & 93.8/{66.2} & 92.7/{65.6} & 94.9/{68.5} & 93.3/{69.7} & 94.5/{68.8} & 94.2/{89.0} & 95.1/{69.3} \\
        & Input-Aware & 94.3/{56.2} & 94.1/{61.1} & 91.7/{60.3} & 92.8/{62.0} & 93.9/{53.4} & 91.1/{52.7} & 94.5/{93.1} & 92.5/{62.2} \\
        & Narcissus & 96.5/{40.7} & 95.0/{60.4} & 93.8/{52.1} & 94.7/{62.7} & 95.9/{53.0} & 95.5/{84.1} & 95.2/{64.5} & 94.3/{61.0} \\
        & LIRA & 95.2/{59.3} & 94.1/{90.3} & 93.1/{68.2} & 94.8/{68.9} & 93.9/{59.7} & 93.7/{69.4} & 94.5/{50.0} & 95.0/{68.6} \\
        & BppAttack & {96.8/{40.6}} & 96.1/{62.0} & 95.0/{51.5} & 96.5/{52.3} & 95.6/{63.7} & 95.9/{83.3} & 95.2/{52.9} & 97.0/{62.6} \\
        & \textbf{BATOD \small(Ours)} & \textbf{97.9/{2.7}} & \textbf{96.8/{10.1}} & \textbf{96.1/{6.0}} & \textbf{97.2/{11.4}} & \textbf{98.1/{8.8}} & \textbf{97.6/{13.7}} & 
        \textbf{97.0/{14.5} } &
        \textbf{96.5/{7.9}} \\
        
        \specialrule{1.5pt}{0pt}{0pt} 
        \multirow{10}{*}{\textbf{ADNI}}
        & BadNets & 92.3/{50.8} & 90.3/{68.0} & 90.0/{57.3} & 91.8/{68.6} & 91.2/{68.4} & 90.9/{59.1} & 92.0/{89.5} & 91.5/{69.9} \\
        & Blended & 93.1/{40.4} & 92.5/{63.0} & 91.3/{53.5} & 92.9/{63.8} & 91.6/{54.4} & 92.1/{64.7} & 91.8/{85.1} & 93.3/{54.1} \\
        & SIG & 94.9/{40.2} & 93.5/{53.8} & 93.0/{62.5} & 94.3/{52.9} & 93.8/{54.3} & 93.2/{84.7} & 94.0/{64.0} & 94.7/{63.2} \\
        & Wanet & 92.2/{52.3} & 91.8/{57.0} & 90.6/{86.9} & 92.0/{57.8} & 91.2/{58.3} & 90.9/{67.6} & 91.5/{68.0} & 92.3/{67.3} \\
        & SSBA & 92.6/{46.7} & 91.0/{55.1} & 90.6/{54.9} & 92.1/{65.4} & 90.9/{66.5} & 91.8/{65.8} & 91.4/{86.0} & 92.5/{56.2} \\
        & Input-Aware & 91.6/{55.8} & 91.4/{57.8} & 90.3/{57.5} & 90.9/{58.1} & 91.1/{69.1} & 90.0/{48.6} & 91.7/{88.9} & 90.6/{68.4} \\
        & Narcissus & 93.7/{39.3} & 93.0/{50.9} & 92.3/{52.0} & 92.8/{62.2} & 93.8/{52.5} & 93.5/{82.8} & 93.2/{63.0} & 92.6/{61.7} \\
        & LIRA & 92.5/{47.9} & 91.6/{87.6} & 90.8/{65.7} & 92.1/{66.5} & 91.4/{57.0} & 91.0/{56.8} & 91.9/{67.3} & 92.4/{66.1} \\
        & BppAttack & 94.1/{39.1} & 93.5/{59.5} & 92.0/{69.1} & 93.8/{69.8} & 92.9/{51.0} & 92.4/{60.3} & 93.1/{82.7} & 94.0/{50.0} \\
        & \textbf{BATOD \small(Ours)} & \textbf{95.3/{1.9}} & \textbf{93.7/{10.1}} & \textbf{93.1/{8.2}} & \textbf{94.3/{9.9}} & \textbf{95.0/{12.3}} & \textbf{94.8/{14.5}} & \textbf{94.0/{9.8}} & \textbf{93.4/{9.0}} \\
        \specialrule{1.5pt}{0pt}{0pt}
        \end{tabular}}
    \label{tab:comparison_defense_attacks_auc}
\end{table*}

\renewcommand{\arraystretch}{2}

\begin{table*}[ht]
    \small 
    \caption{Evaluation of Our Attack's Efficacy Against Various Defense Mechanisms in Out-of-Distribution (OOD) Detection Tasks. An optimal backdoor attack for detecting outliers is quantified by improvements in \textbf{Benign-AUC} (\(\uparrow\)), indicating superior performance, and reductions in \textbf{Poisoned-AUC} (\(\downarrow\)), reflecting decreased susceptibility. Higher \textbf{Benign-AUC} values are advantageous, whereas lower \textbf{Poisoned-AUC} values are favorable. For additional information on inlier and outlier datasets, refer to \ref{appendix:others_dataset}.}

    \vspace{5pt}
    \setlength{\aboverulesep}{1pt}
    \setlength{\belowrulesep}{1pt}
    \renewcommand{\arraystretch}{1.5} 
    \resizebox{\linewidth}{!}{
        \begin{tabular}{@{}ccccccccccc@{}} 
        \specialrule{1pt}{0pt}{0pt} 
        \multirow{2}{*}{\textbf{In-Dataset}} & \multirow{2}{*}{\textbf{Out-Dataset}} & \multirow{2}{*}{\textbf{Attacks}} & \multirow{2}{*}{\textbf{No Defense}} & \multicolumn{7}{c}{\textbf{Defenses \small\textbf{({Benign-AUC} / Poison-AUC)}}} \\
        \cmidrule(lr){5-11}
        & & & & NAD & ABL & ANP & SAU & I-BAU & NPD & RNP \\        
        \specialrule{0.25pt}{0pt}{0pt} 
        \multirow{10}{*}{\textbf{CIFAR10}}
        & \multirow{10}{*}{\textbf{$\text{Others}^*$}} & BadNets & 81.6/{60.3} & 88.7/{67.6} & 81.2/{59.1} & 84.3/{64.1} & 89.7/{68.8} & 80.2/{58.1} & 83.2/{77.6} & 85.2/{65.0} \\
        & & Blended & 85.1/{53.2} & 82.2/{53.2} & \textbf{85.2}/{57.5} & 86.6/{51.0} & 86.0/{61.0} & 87.4/{54.1} & 80.1/{75.7} & 85.2/{55.0} \\
        & & SIG & 88.9/{50.4} & 89.9/{64.8} & 82.8/{45.3} & 87.7/{55.8} & 83.3/{58.0} & 86.4/{82.1} & 84.6/{59.8} & 81.0/{43.1} \\
        &  & Wanet & 80.8/{64.8} & 82.6/{65.1} & 81.9/{75.4} & 85.6/{66.5} & 79.2/{57.0} & 78.9/{58.1} & 78.2/{57.2} & 73.9/{47.6} \\
        &  & SSBA & 83.9/{55.1} & \textbf{91.6}/{60.3} & 80.2/{52.6} & 86.4/{58.0} & 90.3/{60.0} & 87.8/{56.3} & \textbf{88.9}/{84.1} & 83.0/{57.6} \\
        &  & Input-Aware & 79.7/{68.4} & 73.1/{51.6} & 73.4/{56.5} & 80.3{66.9} & 81.3/{51.8} & 89.4/{56.0} & 78.2/{75.5} & 81.0/{62.2}  \\
        &  & Narcissus & 89.1/{48.3} & 83.2/{59.4} & 82.1/{58.0} & 79.2/{47.2} & 85.3/{54.5} & \textbf{91.1}/{90.5} & 88.5/{50.2}  & \textbf{87.7}/{54.1} \\
        &  & LIRA & 82.9/{56.4} & 81.6/{79.1} & 78.9/{56.6} & 70.2/{47.2} & 88.3/{55.4} & 79.9/{56.1} & 78.6/{57.2} & 79.0/{54.6} \\ 
        &  & BppAttack & 90.1/{46.8} & 90.3/{59.1} & 83.4/{67.1} & 83.0/{57.2} & \textbf{91.5}/{64.5} & 90.3/{59.9} & 87.7/{82.1} & 84.2/{58.1} \\ 
        & &  \textbf{BATOD \small(Ours)}  & \textbf{90.3/{7.3}} &  86.6/\textbf{{8.0}} &   80.2/\textbf{{6.1}} &  \textbf{88.6/{7.2}} &  82.0/\textbf{{13.6}} & 88.4/\textbf{{15.8}} & 85.0/\textbf{{10.1}} & 86.4/\textbf{{12.7}} \\
        \specialrule{1pt}{0pt}{0pt} 
        \multirow{10}{*}{\textbf{CIFAR100}}
        & \multirow{10}{*}{\textbf{$\text{Others}^*$}} & BadNets & 79.2 /{59.6} & 88.7/{67.6} & 81.2/{59.1} & 84.3/{64.1} & 89.7/{68.8} & 89.2/{58.1} & 83.2/{77.6} & 85.2/{55.0} \\
        &  & Blended & 85.8{50.1} & 82.2/{53.2} & \textbf{85.2}/{54.5} & 86.6/{57.0} & 86.0/{51.0} & 87.4/{53.1} & 80.1/{75.7} & 85.2/{65.0} \\
        &  & SIG & 87.9/{44.4} & 80.9/{54.8} & 82.8/{55.3} & 87.7/{45.8} & 83.3/{58.0} & 82.4/{76.1} & 84.6/{59.8} & 81.0/{53.1} \\
        &  & Wanet & 77.8/{61.8} & 72.6/{55.1} & 75.9/{69.4} & 75.6/{56.5} & 79.2/{57.0} & 78.9/{57.1} & 78.2/{57.2} & 73.9/{47.6} \\
        &  & SSBA & 82.9/{53.1} & 81.6/{50.3} & 80.2/{54.6} & 86.4/{68.0} & 80.3/{60.0} & 87.8/{54.3} & 85.9/{76.1} & 83.0/{57.6} \\
        &  & Input-Aware & 75.7/{67.4} & 73.1/{51.6} & 78.4/{55.5} & 70.3/{56.9} & 74.7/{51.8} & 79.4/{56.0} & 78.2/{74.5} & 71.0/{42.2}\\
        &  & Narcissus & 89.1/{43.3} & 83.2/{59.4} & 82.1/{68.0} & 79.2/{57.2} & 85.3/{64.5} & \textbf{91.1}/{90.5} & \textbf{88.5}/{60.2}  & \textbf{87.7}/{54.1}\\
        &  & LIRA & 80.9/{55.4} & 81.6/{80.1} & 78.9/{56.6} & 80.2/{57.2} & 78.3/{55.4} & 79.9/{56.1} & 78.6/{57.2} & 79.0/{44.6} \\ 
        &  & BppAttack & 90.0/{36.8} & \textbf{90.3}/{59.1} & 83.4/{67.1} & 83.0/{57.2} & \textbf{91.5}/{64.5} & 90.3/{59.9} & 87.7/{77.2} & 84.2/{68.1}\\ 
        &  & \textbf{BATOD \small(Ours)}  & \textbf{90.5/{9.3}} &  86.6/\textbf{{9.0}} &   80.2/\textbf{{6.7}} &  \textbf{88.3/{7.9}} &  82.0/\textbf{{14.5}} & 88.4/\textbf{{16.3}} & 85.0/\textbf{{11.2}} & 86.4/\textbf{{13.9}} \\
        \specialrule{1.5pt}{0pt}{0pt} 
        
        \end{tabular}}
    \label{tab:osr_auc}
    \footnotesize{\scriptsize $^*$See appendix \ref{appendix:others_dataset}.}
\end{table*}

\begin{table*}[ht]
    \centering
    \caption{Comparative Analysis of Attack Stability Against Various Defensive Measures Employing Two Metrics: \textbf{Benign-ACC}, assessing classification accuracy on benign datasets, and \textbf{Poisoned-ACC}, measuring classification accuracy on datasets compromised by poisoning.}

    \vspace{5pt}
    \setlength{\aboverulesep}{1pt}
    \setlength{\belowrulesep}{1pt}
    \renewcommand{\arraystretch}{1.5} 
    \resizebox{\linewidth}{!}{
        \begin{tabular}{@{}cccccccccc@{}} 
        \specialrule{1pt}{0pt}{0pt} 
        \multirow{2}{*}{\textbf{Dataset}} & \multirow{2}{*}{\textbf{Attack}} & \multirow{2}{*}{\textbf{No Defense}} & \multicolumn{7}{c}{\textbf{Defenses \small\textbf{({Benign-ACC} / Poison-ACC)}}} \\
        \cmidrule(lr){4-10}
        & & & NAD & ABL & ANP & SAU & I-BAU & NPD & RNP \\        
        \specialrule{0.25pt}{0pt}{0pt} 
        \multirow{10}{*}{\textbf{Cityscapes}}
        & BadNets & 91.2 /{11.6} & 88.7/{87.6} & 81.2/{89.1} & 84.3/{84.1} & 89.7/{88.8} & 89.2/{88.1} & 83.2/{57.6} & 85.2/{85.0} \\
        & Blended & 93.5/{0.1} & 92.2/{3.2} & 75.2/{67.5} & 86.6/{61.0} & 86.0/{31.0} & 87.4/{34.1} & 80.1/{45.7} & 85.2/{85.0} \\
        & SIG & 83.9/{4.4} & 90.9/{64.8} & 52.8/{55.3} & 77.7/{45.8} & 83.3/{58.0} & 82.4/{54.1} & 54.6/{89.8} & 71.0/{53.1} \\
        & Wanet & 91.8/{9.8} & 92.6/{75.1} & 81.9/{85.4} & 85.6/{86.5} & 89.2/{87.0} & 88.9/{87.1} & 68.2/{67.2} & 73.9/{47.6} \\
        & SSBA & 92.9/{3.1} & 91.6/{20.3} & 80.2/{84.6} & 86.4/{78.0} & 90.3/{80.0} & 87.8/{74.3} & 68.9/{46.1} & 73.0/{57.6}\\
        & Input-Aware & 89.7/{2.4} & 93.1/{91.6} & 63.4/{5.5} & 90.3{86.9} & 81.3/{61.8} & 89.4/{56.0} & 68.2/{49.5} & 91.0/{92.2} \\
        & Narcissus & 89.1/{3.3} & 83.2/{79.4} & 82.1/{88.0} & 79.2/{37.2} & 85.3/{84.5} & 91.1/{90.5} & 58.5/{20.2}  & 77.7/{34.1}\\
        & LIRA & 72.9/{32.4} & 91.6/{81.1} & 78.9/{66.6} & 20.2/{7.2} & 88.3/{85.4} & 89.9/{86.1} & 78.6/{37.2} & 19.0/{4.6}\\ 
        & BppAttack & 90.2/{6.8} & 90.3/{79.1} & 83.4/{87.1} & 83.0/{87.2} & 91.5/{84.5} & 90.3/{89.9} & 67.7/{47.2} & 84.2/{68.1} \\ 
        & \textbf{BATOD \small(Ours)}  & 90.5/{79.3} &  86.6/{81.2} &   80.2/{71.7} &  88.3/{71.9} &  82.0/{71.9} & 88.4/{87.3} & 75.0/{71.2} & 86.4/{83.9} \\
        \specialrule{1pt}{0pt}{0pt} 
        
        \end{tabular}}
    \label{tab:comparison_defense_attacks}
\end{table*}

\renewcommand{\arraystretch}{2}

\begin{table*}[ht]
    \small 
    \caption{Comparison of the two proposed triggers against various defenses on \textbf{Cityscapes} dataset, with each cell containing four numbers. The right number in each set shows robust OOD detection performance against backdoor attacks. The mean score against all defenses for each attack is provided in the last column. A better backdoor attack for OOD detection is characterized by: Benign-ACC \(\uparrow\), Poisoned-AUC \(\uparrow\). Here, \(\uparrow\) indicates that higher values are better, and \(\downarrow\) denotes that lower values are preferable.}

    \vspace{5pt}
    \setlength{\aboverulesep}{1pt}
    \setlength{\belowrulesep}{1pt}
    \renewcommand{\arraystretch}{1.5} 
    \resizebox{\linewidth}{!}{
        \begin{tabular}{@{}ccccccccccc@{}} 
        \specialrule{1pt}{0pt}{0pt} 
        \multirow{2}{*}{\textbf{Dataset}} & \multirow{2}{*}{\textbf{Attack}} & \multirow{2}{*}{\textbf{No Defense}} & \multicolumn{7}{c}{\textbf{Defenses \small\textbf{({Benign-ACC} / Poison-ACC)}}} \\
        \cmidrule(lr){4-10}
        & & & NAD & ABL & ANP & SAU & I-BAU & NPD & RNP & $\text{AT}^*$ \\        
        \specialrule{0.25pt}{0pt}{0pt} 
        \multirow{3}{*}{\textbf{Cityscapes}}
        & In-Triggers & 86.3/{19.8} & 85.8/{20.0} & 84.9/{24.6} & 85.4/{22.3} & 84.1/{25.3} & 86.0/{25.9} & 85.2/{26.6} & 82.9/{25.2} & 83.0/{27.6} \\
        & Out-Triggers & 85.1/{8.4} & 82.8/{9.8} & 81.4/{9.5} & 83.4/{8.1} & 82.8/{10.2} & 84.0/{12.8} & 83.1/{19.5} & 82.0/{17.6} & 81.1/{21.6} \\
        & \textbf{BATOD \small(Ours)} & 84.9/{2.1} & 83.5/{6.4} & 82.8/{8.3} & 84.7/{5.7} & 83.6/{9.0} & 84.2/{13.4} & 84.6/{12.8} & 83.3/{11.7} & 83.0/{14.6} \\

        \specialrule{1.5pt}{0pt}{0pt} 
        
        \end{tabular}}
    \label{tab:different_trigger}
    
\end{table*}
\begin{table*}[ht]
    \caption{
    An ablation study on our method with the exclusion of different components while keeping the others intact. The result is on the \textbf{Cityscapes} dataset.}

    \vspace{5pt}
    \setlength{\aboverulesep}{1pt}
    \setlength{\belowrulesep}{1pt}
    \renewcommand{\arraystretch}{1.5}
    \resizebox{ \linewidth}{!}
    {\begin{tabular}{@{}ccccc c c c ccccccccc} 

    \specialrule{1.5pt}{\aboverulesep}{\belowrulesep}
    \multirow{3}{*}{\textbf{Setting}} & \multicolumn{2}{c}{\textbf{Trigger Selection}}  &\multicolumn{2}{c}{\textbf{Trigger Type}}  & &\multirow{4}{*}{\textbf{No Defense}} & & \multicolumn{6}{c}{\textbf{Defenses \small\textbf{({Benign-ACC} / Poison-ACC)}}}\\
      \cmidrule(lr{0.25pt}){2-3} 
      \cmidrule(lr{0.25pt}){4-5} \cmidrule(l{0pt}r{0pt}){8-14}


    & \multirow{2}{*}{Random} & \multirow{2}{*}{$\text{Smart}^+$} &  \multirow{2}{*}{In-Triggers} &\multirow{2}{*}{Out-Triggers} & & & \multirow{2}{*}{NAD} & \multirow{2}{*}{ABL} & \multirow{2}{*}{ANP} & \multirow{2}{*}{SAU} & \multirow{2}{*}{I-BAU} & \multirow{2}{*}{NPD} & \multirow{2}{*}{RNP} & \multirow{2}{*}{\textbf{$\text{AT}^*$}}\\

    & \small(without a discriminator) & \small(via a discriminator) & & & & & & & & & & & \\

    \specialrule{1.5pt}{\aboverulesep}{\belowrulesep} 
\textbf{Setup A} & \checkmark & - & \checkmark & - &  \vline & {82.7 /} 3.6  & {82.3 / }11.1 & {81.0 /} 14.6 & {82.6 /} 15.1 & {83.0 / }12.4 & {82.8 /} 17.1 & {82.7 /} 15.6 & {81.9 /} 13.1 & {83.6 /} 22.5 \\
\textbf{Setup B} & \checkmark & - & - & \checkmark & \vline & {81.9 / }4.4 & {81.7 / }10.9 & {80.7 / }15.1 & {81.9 / }15.5 & {82.7 / }12.9 &
{82.4 / }17.5 & 
{82.3 / }15.8 & 
{81.5 / }13.4 & {83.2 /} 23.4 \\
\textbf{Setup C} & \checkmark & - & \checkmark & \checkmark & \vline & {83.1 / }2.9 & {82.9 / }10.5 & {81.3 / }14.2 & {83.0 / }14.8 & {83.2 / }12.0 & {83.1 /} 16.7 & {83.0 /} 15.1 & {82.1 /} 12.6 & {84.1 /} 21.7  \\
\textbf{Setup D} & - & \checkmark & \checkmark & - & \vline & {83.8 / }2.5 & {83.4 / }9.7 & {81.9 / }13.6 & {83.8 / }14.2 & {84.2 / }11.5 & {83.7 /} 16.1 & {83.6 /} 14.5 & {82.9 /} 12.0 & {84.9 /} 19.7 \\

\textbf{Setup E} & - & \checkmark & - & \checkmark & \vline &  {83.5 /} 2.8 & {83.2 /}10.0  & {81.6 /} 14.0 & {83.4 /} 14.5 &  {83.9 /} 11.7 & {83.5 /} 16.4 & {83.4 /} 14.7 & {82.4 /} 12.3 & {84.7 /} 20.3 \\

    \textbf{Setup F \small{(Ours)}} \ & - & \checkmark & \checkmark & \checkmark & \vline &  \textbf{84.1/{2.3}} & \textbf{83.6/{9.2}} & \textbf{82.2/{13.2}} & \textbf{84.0/{13.9}} & \textbf{84.5/{11.1} } & \textbf{84.2/{15.8}} & \textbf{83.9/{14.2}} & \textbf{83.5/{11.7}} &
    \textbf{85.1/19.1}\\

    \noalign{\vskip 2pt}

    \specialrule{1.5pt}{\aboverulesep}{\belowrulesep}

     \end{tabular}}
    \label{tab:setups_table}
    \small\small
    \footnotesize{{\tiny $^+$ Selecting triggers via $f_s$ to embed in the data points.\\
    $^*$ In this table we use Adversarial Training (AT) techniques to create robust models.}
}
\end{table*}
\renewcommand{\arraystretch}{2}

\begin{table*}[ht]
\small
\caption{Comparison of different OOD scoring functions against various defenses on \textbf{Cityscapes} as in-distribution and \textbf{PreAct-ResNet18} as model, with each cell containing two numbers. The mean score for each scoring function is provided in the last column.}
\vspace{5pt}
\setlength{\aboverulesep}{1pt}
\setlength{\belowrulesep}{1pt}
\renewcommand{\arraystretch}{1.5}
\scriptsize
\resizebox{\linewidth}{!}{
    \begin{tabular}{@{}ccccccccccc@{}}
    \specialrule{1pt}{0pt}{0pt} 
        \multirow{2}{*}{\textbf{Dataset}} & \multirow{2}{*}{\textbf{$\text{SF}^*$}} & \multirow{2}{*}{\textbf{No Defense}} & \multicolumn{7}{c}{\textbf{Defenses \small\textbf{({Benign-AUC} / Poison-AUC)}}} \\
        \cmidrule(lr){4-10}
        & & & NAD & ABL & ANP & SAU & I-BAU & NPD & RNP \\        
        \specialrule{0.25pt}{0pt}{0pt} 
    \multirow{4}{*}{\textbf{Cityscapes}} &
    \textbf{MSP} & \textbf{84.1/{2.3}} & \textbf{83.6/{9.2}} & \textbf{82.2/{13.2}} & \textbf{84.0/{13.9}} & \textbf{84.5/{11.1}} & \textbf{84.2/{15.8}} & \textbf{83.9/{14.2}} & \textbf{83.5/{11.7}} \\
    & ODIN & 79.7/{7.3} & 78.3/{14.2} & 76.1/{18.6} & 78.8/{17.9} & 79.2/{16.0} & 79.0/{19.7} & 78.0/{19.5} & 78.3/{17.0} \\
    & KNN & 83.9/{2.5} & 83.2/{9.3} & 82.0/{13.7} & 83.8/{14.1} & 84.1/{11.5} & 84.0/{16.0} & 83.6/{14.3} & 83.4/{11.9} \\
    & GMM & 80.2/{5.1} & 79.8/{12.4} & 77.6/{16.5} & 80.1/{16.6} & 80.6/{14.7} & 80.3/{18.2} & 79.9/{17.3} & 79.6/{15.1} \\
    \specialrule{1.5pt}{0pt}{0pt} 
    \end{tabular}
}
\label{tab:scoring_function}
\footnotesize $^*$Scoring Functions (SF).
\end{table*}


    

\begin{figure}[ht]
\centering
\includegraphics[width=\linewidth]{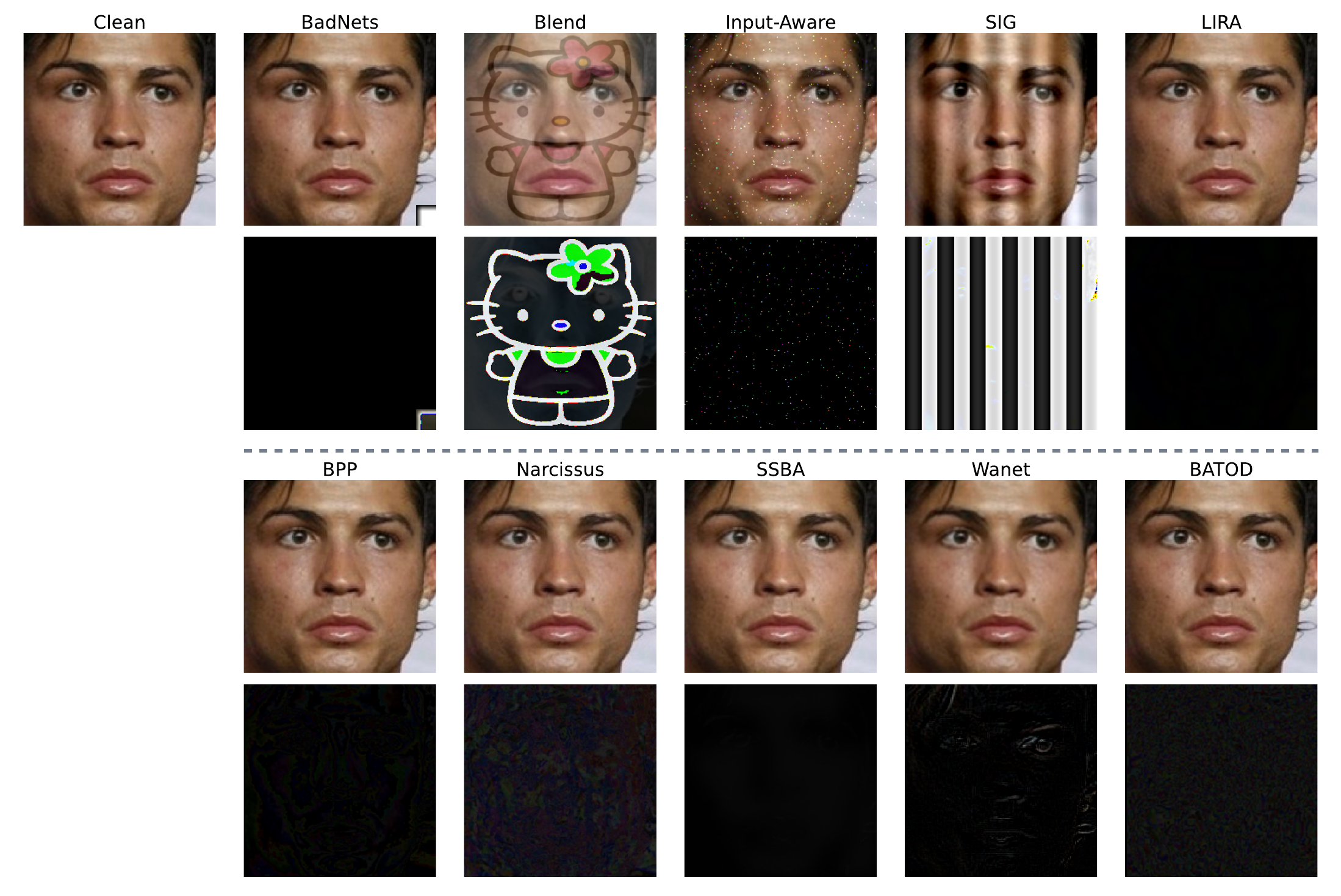}  
\caption{Comparison of backdoor attacks: In the top row, you can view trojaned samples for each attack, and in the bottom row, the residual image is calculated by subtracting the clean sample from the trojaned sample.}
\label{fig:visualize_attacks_completed}

\end{figure}

\section{Experimental Evaluation}
\label{sec:experiment}

\textbf{Experimental Setup and Datasets.} We utilized a diverse array of datasets to evaluate the performance of outlier detection in different scenarios, specifically focusing on open-set recognition (OSR) and out-of-distribution (OOD) detection. For the OSR task, we employed the Cityscapes \cite{Cordts2016Cityscapes}, ADNI \cite{ADNI_data}, and PubFig \cite{kumar2009attribute} datasets. The Cityscapes dataset was used to differentiate between common urban elements and uncommon objects in urban scenes. The ADNI MRI dataset helped us identify signs of Alzheimer's disease against normal cognitive aging. Meanwhile, the PubFig dataset allowed us to assess facial recognition systems' ability to distinguish between known and unknown public figures.\\
For the OOD detection task, we used CIFAR-10 or CIFAR-100 as in-distribution samples and evaluated their performance against OOD samples from LSUN, BIRDS, FOOD, Flowers, and the remaining CIFAR dataset not used as in-distribution. We show the results of the OOD detection task in table \ref{tab:osr_auc} on the named out-distribution dataset as \textbf{others}. For detailed information on the settings and specifics of each dataset, see the Appendix \ref{appendix:others_dataset}.

\textbf{Implementation Details.} We use the ResNet18 as the surrogate model and the PreAct-ResNet18 as our victim model. Also, running each experiment on the GeForce RTX 3090 Ti. The execution time for each attack is between 1100 to 3600 seconds and for defenses is about 2100 seconds. More details about the model architectures and hyperparameters during training and testing exist in Appendix \ref{apendix:implementation_detail}.
 Results from employing different victim models are detailed in Section \ref{sec:ablation_model}.

\textbf{Evaluation Details.}
To do the evaluation, we mapped our triggers on the input datasets via a simple pre-trained discriminator. The evaluation of our model's performance against various defenses using the Cityscapes, Pubfig, and ADNI datasets, is illustrated in Table \ref{tab:comparison_defense_attacks_auc}. To facilitate comparison, we report the Poisoned-AUC, applicable when the input data contains a trigger, and the Benign-AUC, applicable when the input data is free of triggers. These metrics are employed to assess the efficacy of each attack within the robust OSR task.
Furthermore, to demonstrate the effectiveness of our proposed attack in OOD detection tasks, we consider CIFAR10 and other datasets (e.g. CIFAR100, LSUN, etc.) and another did the task on CIFAR100 as the inlier dataset and the others as the outliers. The outcomes of these evaluations are presented in Table \ref{tab:osr_auc}.
Moreover, as evidenced by the data in Table \ref{tab:comparison_defense_attacks}, the BATOD method, distinct from other attacks, specifically avoids targeting the classification accuracy and does not disrupt it. See Appendix \ref{appendix:detailed results} for more information.

\textbf{Analyzing Results}. Our analysis evaluates various attack strategies against defense mechanisms across multiple datasets, as depicted in Tables \ref{tab:comparison_defense_attacks_auc}, \ref{tab:osr_auc}, and \ref{tab:comparison_defense_attacks}. Table \ref{tab:comparison_defense_attacks_auc}'s focus on OOD detection across the Cityscapes, Pubfig, and ADNI datasets reveals significant performance differentials using Benign-AUC and Poisoned-AUC metrics, highlighting the effectiveness of our BATOD method. This method distinctly maintains high Benign-ACC and controlled Poisoned-ACC, demonstrating robustness without compromising classification accuracy. In contrast, other attacks like BadNets and Blended show varied effectiveness under defenses like NAD, ABL, and ANP. Table \ref{tab:osr_auc} further validates our attack's efficacy in OOD tasks using benchmarks such as CIFAR10 and CIFAR100, emphasizing the need for continuous refinement of attack methods and defensive strategies to ensure enhanced security in practical settings. This comprehensive evaluation stresses the importance of ongoing development and adaptation of security measures to maintain robustness in real-world applications.

\section{Ablation}
\label{sec:ablation}

In this section, we initially demonstrate the significance of each component of our method, followed by the application of additional post-hoc scoring functions to further validate the robustness of BATOD (extra ablation is in Appendix \ref{Appendix_Additional_Ablation})

\textbf{Method Components}. In our comprehensive analysis of a proposed backdoor attack strategy, we explore the differential impact of two uniquely designed trigger types: in-triggers and out-triggers, each tailored to subtly manipulate classifier behavior under normal and anomalous conditions, respectively. An ablation study, detailed in Table \ref{tab:setups_table}, assesses the effectiveness of these triggers individually and in combination across various defense scenarios—ranging from no defense to sophisticated backdoor defenses and adversarial training reflecting Trojai competition rounds 3 and 4 conditions. This evaluation, conducted on the diverse Cityscapes dataset, aims to establish baseline performance and test the robustness of the model under real-world conditions. The findings are pivotal in identifying defense vulnerabilities and refining trigger mechanisms to optimize attack efficacy and subtlety.

\textbf{Post-hoc scoring methods}.
\label{sec:ablation_score_function}
To assess the effectiveness of our backdoor attack method across diverse scenarios, we utilized a variety of out-of-distribution (OOD) scoring functions, each offering unique insights into how well the method performs under different detection strategies. MSP (Maximum Softmax Probability) utilizes the highest softmax output of a neural network as the score to distinguish between in-distribution and OOD samples.
ODIN (Out-of-distribution detector for Neural networks) enhances OOD detection by applying temperature scaling and input perturbation to the softmax scores. KNN (K-nearest neighbours) detects OOD samples by measuring the distance to the K nearest neighbours in the feature space, with farther distances indicating OOD. GMM (Gaussian Mixture Model) uses the likelihood of a sample under a fitted Gaussian mixture model to determine if it's OOD, with lower likelihoods signalling out-of-distribution instances. Each method contributes a unique perspective to OOD detection, allowing for a comprehensive evaluation of the backdoor attack's robustness against various detection techniques. The table \ref{tab:scoring_function} shows the performance of our backdoor attack method considering different OOD scoring functions.

\section{Conclusion}
Our work presents a novel backdoor attack methodology that effectively bypasses contemporary out-of-distribution (OOD) detection techniques and remains resilient against various backdoor defenses. By meticulously crafting triggers and employing sophisticated synthesis of out-of-distribution samples, our approach demonstrates its robustness across different datasets, model architectures, and defense mechanisms. Our experimental results underline the critical need for advancing current defense strategies to counteract such advanced backdoor threats. This research not only sheds light on the evolving landscape of backdoor attacks but also prompts a re-evaluation of defense mechanisms in the AI security domain. Future work should focus on developing more sophisticated detection and defense mechanisms that can adapt to the ever-evolving techniques of backdoor attacks, ensuring the integrity and reliability of machine learning models in real-world applications.

\newpage

\bibliographystyle{unsrt}
\bibliography{template}

\newpage
\maketitle
\section*{Appendix}

\appendix
\setcounter{secnumdepth}{2}

\label{appendix}


\section{Detailed Results} 
\label{appendix:detailed results}

In this Section, in Tables \ref{tab:std_results}, we provide the mean and standard deviation of our method's results on the provided datasets in Table \ref{tab:comparison_defense_attacks_auc} using 10 different seeds just on Cityscapes dataset. These were not reported in the main table due to space constraints.

\renewcommand{\arraystretch}{2}

\begin{table*}[ht]
    \centering
    \caption{Detailed results of our method’s performance on the Cityscapes dataset, over 10 runs}

    \vspace{5pt}
    \setlength{\aboverulesep}{1pt}
    \setlength{\belowrulesep}{1pt}
    \renewcommand{\arraystretch}{1.5} 
    \resizebox{\linewidth}{!}{
        \begin{tabular}{@{}cccccccc@{}} 
        \specialrule{1pt}{0pt}{0pt} 
        \noalign{\smallskip}
         \multirow{2}{*}{\textbf{No Defense}} & \multicolumn{7}{c}{\textbf{Defenses \small\textbf{({Benign-AUC} / Poison-AUC)}}} \\
        \cmidrule(lr){2-8}
        & NAD & ABL & ANP & SAU & I-BAU & NPD & RNP \\        
        \specialrule{0.25pt}{0pt}{0pt} 
         \textbf{84.1 $\pm$ 0.8 / {2.3 $\pm$ 0.1}} & \textbf{83.6 $\pm$ 0.6 / {9.2 $\pm$ 0.3}} & \textbf{82.2 $\pm$ 0.8 / {13.2 $\pm$ 0.5}} & \textbf{84.0 $\pm$ 0.1 / {13.9 $\pm$ 0.7}} & \textbf{84.5 $\pm$ 0.2 / {11.1 $\pm$ 0.6} } & \textbf{84.2 $\pm$ 0.5 / {15.8} $\pm$ 0.2} & \textbf{83.9 $\pm$ 0.9 / {14.2} $\pm$ 0.7 } & \textbf{83.5 $\pm$ 0.3 / {11.7} $\pm$ 0.1 } \\
        \specialrule{1.5pt}{0pt}{0pt}
        \end{tabular}}
    \label{tab:std_results}
\end{table*}

\section{Datasets}
\label{Appendix_Dataset_Details}

\subsection{Cityscapes} The Cityscapes dataset, meticulously assembled for facilitating advancements in semantic urban scene understanding, comprises a substantial corpus of over 5,000 finely annotated images and 20,000 coarsely annotated images, captured across 50 diverse cities. This dataset, designed with a focus on instance-level semantic labelling and rich annotations, includes varied annotations captured under controlled weather conditions using a sophisticated stereo camera setup. Such comprehensive data, including pixel-level and instance-level annotations for dynamic and static elements like vehicles and pedestrians, proves invaluable in developing advanced vision models tailored for complex urban scenarios, particularly autonomous driving.

\begin{figure}[h]
 \centering
\begin{minipage}{.5\textwidth}

    \centering
    \begin{subfigure}[b]{0.5\textwidth}
       \centering 
       \includegraphics[width=\linewidth]{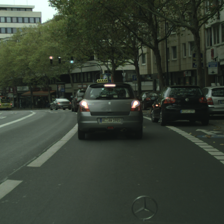}
       \caption{An inlier sample}
       \label{fig:strip_cifar10}
    \end{subfigure}%
    \begin{subfigure}[b]{0.5\textwidth}
       \centering 
       \includegraphics[width=\linewidth]{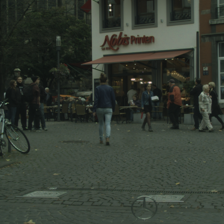}
       \caption{An outlier sample}
    \end{subfigure}
    \caption{Two sample images from \textbf{Cityscapes} dataset}
      \label{fig:cityscapes}
  
\end{minipage}%
\end{figure}

In our research, we have repurposed the Cityscapes dataset for an outlier detection task, inspired by methodologies such as those employed by OpenGAN. You can see two sample images of this setting in Figure \ref{fig:cityscapes}. By designating less frequent semantic classes, categorized under "other," as outliers, and more frequent urban elements as inliers, we explore the capability of machine learning models to distinguish between typical urban components and anomalous, ambiguous objects. This setup leverages the rich annotations and diverse imagery of Cityscapes to evaluate and enhance the precision with which models recognize and react to atypical features in urban settings. Such an approach is crucial not only for validating the effectiveness of outlier detection Algorithms but also for reinforcing their practical applicability in scenarios critical to advancing autonomous driving technologies, where accurately identifying both common and uncommon elements is vital.
\subsection{ADNI Neuroimaging}

The study focuses on differentiating between early signs of Alzheimer's disease and normal cognitive ageing by classifying the Alzheimer’s Disease Neuroimaging Initiative (ADNI) dataset into six distinct classes based on cognitive status and disease progression. The classes CN (Cognitively Normal) and SMC (Subjective Memory Concerns) are considered inliers, representing individuals without any significant neurodegenerative disease or with only minor memory concerns, respectively. Conversely, the classes AD (Alzheimer's Disease), MCI (Mild Cognitive Impairment), EMCI (Early Mild Cognitive Impairment), and LMCI (Late Mild Cognitive Impairment) are treated as outliers, covering a spectrum from mild to severe cognitive impairments associated with Alzheimer's progression.

\begin{figure}[h]
\centering
\begin{minipage}{.5\textwidth}

    \centering
    \begin{subfigure}[b]{0.5\textwidth}
       \centering 
       \includegraphics[width=\linewidth]{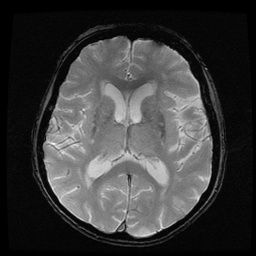}
       \caption{An outlier sample}
    \end{subfigure}%
    \begin{subfigure}[b]{0.5\textwidth}
       \centering 
       \includegraphics[width=\linewidth]{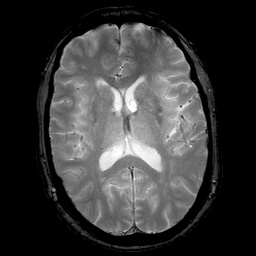}
       \caption{An inlier sample}
    \end{subfigure}
    \caption{Two sample images from \textbf{ADNI} dataset}
      \label{fig:adni}
  
\end{minipage}%
\end{figure}

Each class contains 6000 records, a paired sample shows in Figure \ref{fig:adni}, ensuring a comprehensive representation across different stages of Alzheimer's disease and cognitive health:

\textbf{AD}: Patients diagnosed with Alzheimer’s disease.
\\
\textbf{CN}: Individuals showing no signs of cognitive impairment.
\\
\textbf{MCI}: Represents a transitional stage between normal cognitive ageing and more severe decline.
\\
\textbf{EMCI}: Patients with milder symptoms of cognitive decline.
\\
\textbf{LMCI}: Patients closer to Alzheimer’s disease in terms of symptom severity.
\\
\textbf{SMC}: Individuals who report memory concerns yet perform normally on cognitive tests.

The analysis utilizes the middle slice of MRI scans from ADNI rounds 1, 2, and 3, adhering to the guidelines established by ADNI. These guidelines are designed to standardize imaging procedures and ensure consistency in the data used for research into Alzheimer’s disease. This methodological approach maintains a consistent focus on comparable brain regions across all participants and time points, facilitating reliable longitudinal studies and cross-sectional comparisons within the machine learning framework.

\subsection{Pubfig}
The PubFig dataset, or Public figures Face Database, is a comprehensive resource containing over 58,000 images of 200 celebrities, crafted to enhance research in facial recognition technologies. Each public figure is represented through a diverse array of images that capture a wide range of poses, lighting conditions, and expressions, designed to test the robustness of facial recognition systems under realistic scenarios. In our study, we have tailored this dataset to a specific experimental design by selecting a subset of 50 celebrities. From this subset, we randomly designated 25 celebrities as inliers and the remaining 25 as outliers. This methodology enables us to scrutinize the ability of facial recognition systems to distinguish between known (inlier) and less familiar or potentially deceptive (outlier) faces, thereby evaluating the resilience of these systems against various adversarial attacks and contributing to the advancement of security measures in biometric technologies.

\subsection{Out-of-distribution detection (CIFAR10 and CIFAR100)}
\label{appendix:others_dataset}
Our method is effective across various computer vision tasks, models, and datasets, demonstrating its broad applicability. To illustrate this, we conducted tests using the CIFAR-10 and CIFAR-100 datasets as in-distribution samples, and LSUN, BIRDS, FOOD, Flowers, and one of either CIFAR-10 or CIFAR-100 (which is not considered in-distribution) as OOD samples. To avoid overlap, we removed classes from the OOD datasets that were present in the in-distribution datasets.

\section{Extra Ablations}
\label{Appendix_Additional_Ablation}

\subsection{Model Architecture}
\label{sec:ablation_model}

\renewcommand{\arraystretch}{2}

\begin{table}[ht]
    \small
    \centering
    \caption{Evaluation of the BATOD Method's Efficacy Across Various Model Architectures in Out-of-Distribution (OOD) Detection Tasks, Utilizing CIFAR10 as the In-Distribution and ResNet18 as the Surrogate Model. An effective backdoor attack for OOD detection is defined by a measurable increase in C-AUC (\(\uparrow\)), which suggests improved performance, coupled with a decrease in R-AUC (\(\downarrow\)), implying reduced vulnerability. These metrics provide a dual perspective on the efficacy of the attack: higher values of C-AUC are considered better as they indicate a stronger capability to correctly classify OOD instances, whereas lower values of R-AUC are deemed preferable as they reflect the attack's reduced impact on normal operations. Models marked with an asterisk (*) are pre-trained, indicating they have undergone prior conditioning to enhance their initial performance before being subjected to OOD detection scenarios.}

    \vspace{5pt}
    \setlength{\aboverulesep}{1pt}
    \setlength{\belowrulesep}{1pt}
    \renewcommand{\arraystretch}{1.5}
    \scriptsize
    \resizebox{\linewidth}{!}{
        \begin{tabular}{@{}ccccccccc@{}}
        \specialrule{1pt}{0pt}{0pt} 
        \multirow{2}{*}{\textbf{Model}} & \multirow{2}{*}{\textbf{No Defense}} & \multicolumn{7}{c}{\textbf{Defenses \small{({Benign-ACC} / Poison-ACC)}}} \\
        \cmidrule(lr){3-9}
        & & NAD & ABL & ANP & SAU & I-BAU & NPD & RNP \\        
        \specialrule{0.25pt}{0pt}{0pt}
        \textbf{PreAct-ResNet18} & {90.5/{9.3}} &  86.6/{{9.0}} &   80.2/{{6.7}} &  {88.3/{7.9}} &  82.0/{{14.5}} & 88.4/{{16.3}} & 85.0/{{11.2}} & 86.4/{{13.9}}  \\
        \textbf{ResNet18} & 91.7/{8.4} & 87.1/{8.5} & 81.3/{5.2} & 89.1/{7.1} & 82.3/{14.0} & 89.3/{15.9} & 86.3/{10.8} & 87.3/{12.6} \\
        \textbf{ResNet18*} & 93.6/{6.3} & 87.4/{8.1} & 82.7/{4.9} & 89.7/{6.6} & 87.3/{13.6} & 89.8/{15.0} & 87.1/{10.2} & 88.1/{11.9} \\
        \textbf{Wide-ResNet-40-4} & 92.9/{7.3} & 86.7/{8.7} & 81.5/{5.3} & 89.0/{7.5} & 86.6/{14.1} & 89.1/{15.4} & 86.3/{11.1} & 87.7/{12.3} \\
        \textbf{ViT} & 94.0/{5.2} & 89.1/{7.2} & 85.7/{4.7} & 95.7/{53.1} & 87.9/{13.3} & 90.3/{14.5} & 87.8/{10.4} & 88.9/{11.2}  \\
        \textbf{ViT*} & 95.6/{4.6} & 90.7/{6.0} & 89.2/{4.0} & 96.2/{49.4} & 89.1/{12.8} & 91.6/{13.3} & 88.9/{9.7} & 89.6/{10.0} \\
        \specialrule{1.5pt}{0pt}{0pt} 
        \end{tabular}
    }
    \label{tab:model_architecture}
\end{table}

In this section, we examine our backdoor attack method by applying it to various models, including different model architectures, like Residual Neural Networks \cite{he2016deep}, Wide Residual Networks \cite{zagoruyko2017wide} and Vision Transformers \cite{dosovitskiy2020image}, in addition, we have tested pre-trained models as well. You can see the outlier detection performance of our method on different models in Figure \ref{tab:model_architecture}.

\noindent\textbf{Comparison with More Attacks.}\ \ In \ref{fig:visualize_attacks_completed} you can see triggers for different backdoor attack methods.

\noindent\textbf{Base Implementation.}\ \ We adapted implementation of \cite{wu2022backdoorbench} for our task.


%


\begin{algorithm}
\caption{DeepFool: multi-class case}
\label{alg:deepfool_algorithm}
\begin{algorithmic}[1]
\State \textbf{input:} Image $x$, classifier $f$, estimated label $\hat{k}(x_i)$.
\State \textbf{output:} Perturbation $\hat{r}$.
\State
\State Initialize $x_0 \leftarrow x$, $i \leftarrow 0$.
\While{($\hat{k}(x_i) = \hat{k}(x_0)$ and $MSP(f(x_i)) > \frac{1}{2}$)}
    \For{$k \neq \hat{k}(x_0)$}
        \State $w'_k \leftarrow \nabla f_k(x_i) - \nabla f_{\hat{k}(x_0)}(x_i)$
        \State $f'_k \leftarrow f_k(x_i) - f_{\hat{k}(x_0)}(x_i)$
    \EndFor
    \State $l \leftarrow \arg\min_{k \neq \hat{k}(x_0)} \frac{|f'_k|}{\|w'_k\|_2}$
    \State $r_i \leftarrow \frac{|f'_l|}{\|w'_l\|_2^2} w'_l$
    \State $x_{i+1} \leftarrow x_i + r_i$
    \State $i \leftarrow i + 1$
\EndWhile
\State \textbf{return} $\hat{r} = \sum_i r_{i-1}$
\end{algorithmic}
\end{algorithm}

\begin{algorithm}
\caption{Poisoning in Inference Time}
\begin{algorithmic}[1]
\State \textbf{input:} Surrogate model $f_s$, Available part of train-set $D_a$, Binary classifier for outlier detection $H(x)$, Image $x$, Estimated label $\hat{k}(x_i)$.
\State \textbf{output:} Modified image $x'$ for inference.

\State Select the class with the highest probability based on $f_s(x)$ as estimated label $\hat{k}(x_i)$. 
\If{$H(x)$ == "Inlier"}
    \State Calculate $x' = x + \Delta^{Out}_{\hat{k}(x_i)}$ \Comment{Apply out-trigger to make inlier look like outlier}
\Else
    \State Calculate $x' = x + \Delta^{In}_{\hat{k}(x_i)}$ \Comment{Apply in-trigger to make outlier look like inlier}
\EndIf
\State \Return $x'$
\end{algorithmic}
\end{algorithm}

\section{High-frequency constraint for ensuring stealthiness in frequency domain}
Incorporating the insights derived from our discussions on backdoor attack methodologies and the innovative aspects of our trigger generator, the BATOD method significantly enhances resistance against frequency-based defenses. Our approach leverages a sophisticated two-pronged technique to enhance the effectiveness of triggers against such defenses. Noise is initially generated through a noise-additive function, parameterized by \(\phi\), ensuring the noise exists within the constrained range of \([-1, 1]\). This method effectively camouflages the triggers against defenses by ensuring that the noise adheres to the natural frequency ranges of the image data. Specifically, we apply a 2D Discrete Cosine Transform (DCT) to the noise matrix, retaining only low to mid-range frequency components using the matrix \( m \), where:
\[
m_{i,j} = 
\begin{cases} 
1 & \text{if } 1 \leq i, j \leq rd \\
0 & \text{otherwise} `
\end{cases}
\]
The resulting filtered noise is then integrated using:
\[
Q_{\phi}(x) = \text{IDCT}(m \odot \text{DCT}(g_{\phi}(x)))
\]
Further enhancing robustness, a Gaussian blur filter \( k \) is applied, smoothing transitions and diminishing high-frequency components that are typically detectable by defensive Algorithms:
\[
g_{\phi}(x) = (x + Q_{\phi}(x)) \ast k
\]
This dual approach, focusing on the strategic manipulation of frequency components and the integration of adaptive blending techniques, ensures that the BATOD method effectively circumvents sophisticated frequency-based defenses, maintaining the integrity and intended effects of the triggers without detection.


\section{Preliminaries}
\textbf{Vulnerability of Classifiers to Adversarial Manipulations}
\label{appendix:vulnerabiliy_to_adversarial_attack}
Research has demonstrated that deep neural networks (DNNs) are susceptible to adversarial manipulations, especially within classification contexts. When adversarial disturbances are introduced to the input during the prediction phase, they can lead classifiers to incorrectly assign labels. Specifically, a disturbance $\delta$ is Appended to an instance $x$ with its true label $y$, deceiving the model into predicting an incorrect label \(\hat{y}\) for the manipulated input \(x^{*}\), defined as \(x^{*} = x + \delta\). Among various methods, the Projected Gradient Descent (PGD) technique is notable for its iterative approach to generating adversarial instances by keeping the perturbation within the bounds of the \(\ell_{\infty}\) norm over N iterations:

$$x_{0}^{*} = x, \quad x_{t+1}^{*} = x_{t}^{*} + \alpha \cdot \text{sign}\left(\nabla_x L\left( x_{t}^{*}, y\right)\right), \quad x^{*}=x_{N}^{*}$$

Here, \( L\left( x_{t}^{*}, y \right) \) represents the targeted loss function, typically cross-entropy, for manipulating classifier behaviour.

\section{Related Work}

\textbf{Outlier Detection:} The outlier detection task refers to identifying samples that diverge from those inliers. During training, a dataset $D$, with labels assumed to be ID, is used. In most previous works \cite{kong2021opengan,fort2021exploring,hendrycks2016baseline}, the detector $f$ is trained on the task of classifying ID classes. An outlier detector operates by assigning an outlier score to an input sample, where a higher score corresponds to outlier samples. The main challenge in outlier detection is finding an ideal score function that can assign scores to separate ID and outlier samples effectively. Existing score functions for an input test sample $x$ operate on the logits of $f(x)$, hypothesizing that outlier sample logits differ from ID ones. Typically, outlier samples' logits are more uniform because they are unseen by $f$ \cite{ruff2021unifying,salehi2021unified}.

\textbf{Backdoor Attacks Overview:} Gu et al. \cite{gu2017badnets} were the first to demonstrate a backdoor attack on deep neural network (DNN) models, using pixel modifications as a trigger that, while noticeable to humans, initiated a new era in stealthier attack methodologies. Advances in this area have led to two main strategies for hiding these triggers. Invisible triggers, which are subtle changes hard to detect by eye by minimizing the differences in pixels between altered and original images\cite{li2020invisible, chen2017targeted, zhong2020backdoor}, including techniques that maintain similarity in the model's internal representations of both\cite{doan2021backdoor, zhao2022defeat}; and natural triggers, which subtly alter the image's style to appear normal, employing methods like mimicking natural reflections \cite{liu2020reflection}, using social media-style filters\cite{liu2019abs}, applying generative adversarial networks\cite{cheng2021deep}, or performing image transformations\cite{nguyen2021wanet}. A common factor between all of these attacks is focusing on saving clean classification accuracy while achieving a high attack success rate (ASR), which measures the percentage of predicting attack target labels when adding a trigger to an image from another class, but in this work, we focus on out-of-distribution detection of backdoor attacks, and therefore we use other metrics explained in Figure \ref{sec:experiment}.

\textbf{BadNets:} This seminal work pioneered the study of backdoor attacks in deep learning. The approach involved embedding a small, fixed pattern at a predetermined location within an image. By replacing the original pixels in this manner, a poisoned image was created. This method effectively demonstrated how seemingly innocuous modifications could manipulate the behaviour of neural networks, raising significant concerns about the security of machine learning models.

\textbf{Blended:}  Building upon the foundational work of BadNets, this study advances the concept of stealthier backdoor attacks by utilizing alpha blending to create less detectable triggers. This method integrates the trigger more seamlessly into the image, significantly enhancing the invisibility of the manipulation, and thereby complicating detection.

\textbf{SIG:} This work introduces a backdoor attack that leverages a sinusoidal signal as the trigger. The approach perturbs clean images of a target class without altering their labels, thereby maintaining label consistency while effectively implementing a backdoor attack.

\textbf{Input-aware Dynamic Backdoor Attack (Input-aware):} This method introduces a training-controllable attack that simultaneously learns the model parameters and a trigger generator. During testing, the unique trigger generator produces a distinct trigger for each clean testing sample

\textbf{Sample-specific Backdoor Attack (SSBA):} This approach enhances the complexity of backdoor attacks by utilizing an auto-encoder to integrate a trigger, such as a string, into clean samples, thus generating poisoned samples. The uniqueness of this method lies in its ability to produce a distinct residual between the poisoned and clean samples for each individual image, making the perturbations specific to each sample.

\textbf{Warping-based Poisoned Networks (WaNet):} This research introduces a training-controllable attack utilizing a fixed warping function to subtly distort clean samples. The uniqueness of this method lies in the attacker’s ability to control the training process to ensure that only the designated warping function triggers the backdoor, enhancing the attack's stealthiness and specificity.

\subsection{Backdoor Defenses Overview}
defense strategies for models focus on validating or countering the influence of third-party models before they are put into use. ABS\cite{liu2019abs} approached the problem by inspecting neurons and creating trigger possibilities through a reverse-engineering process, which were then tested on clean images. Zhao et al.\cite{zhao2020bridging} introduced a method that indirectly uses mode connectivity to investigate backdoor threats, effectively reducing their impact while preserving the model's accuracy on clean data. Neo\cite{udeshi2022model} identifies potential backdoor triggers by noting changes in predictions when certain image regions are obscured. These techniques confirm their suspicions by applying the identified triggers to a batch of clean images, relying on the assumption that triggers will consistently reveal themselves.

\textbf{Adversarial Neuron Pruning (ANP)} \cite{wu2021adversarial} is predicated on the observation that in a backdoored model, certain neurons are more susceptible to adversarial perturbation (deliberate adjustment of neuron weights to initiate an adversarial attack) compared to others. Drawing on this insight, ANP advocates for the removal of these vulnerable neurons as a countermeasure against backdoor threats. ANP presupposes that certain neurons are predominantly linked with backdoor triggers, but since our attack does not rely on this principle, it effectively circumvents ANP's defenses. Despite ANP's efficacy against many other attacks, our backdoor strategy successfully bypasses it.

\textbf{Anti-Backdoor Learning (ABL)} \cite{li2021anti} leverages the observable pattern wherein, during the initial epochs of training, the loss associated with poisoned samples decreases more swiftly than that of benign samples. Motivated by this observation, ABL implements a strategy to segregate poisoned samples from benign ones by exploiting the variance in the rate of loss reduction. Following segregation, the approach endeavours to neutralize the backdoor threat by intentionally increasing the loss of the segregated poisoned samples, thereby diminishing their potential impact on the model's learning process. Although ABL excels against many backdoor attacks by identifying fixed patterns, it struggles against diverse triggers informed by a backdoored model's knowledge of the target class. Thus, our proposed backdoor attack method can bypass ABL's defenses.

\section{Resilience to more defenses}
\textbf{Neural Cleanse}, developed by  \cite{wang2019neural}, is a method to protect models by focusing on creating specific patterns. It begins with the premise that the harmful part of the model is akin to a small, added-on patch. For every category, Neural Cleanse determines the optimal pattern to transform any safe input into that category. It then identifies any class with a much smaller pattern, which could indicate a hidden issue. This method measures the problem using the Anomaly Index with a specific cutoff point ($\tau = 2$). We tested our backdoor-attacked models with Neural Cleanse and shared the results in \ref{tab:osr_auc}, \ref{tab:model_architecture}, and \ref{tab:scoring_function}. If NC fails to detect the backdoor attack, the model weights remain unchanged. Our attack performed well and was not detected in various scenarios against this defense.

\begin{figure}[h]
\centering
\begin{minipage}{.5\textwidth}

    \centering
    \begin{subfigure}[b]{0.5\textwidth}
       \centering 
       \includegraphics[width=\linewidth]{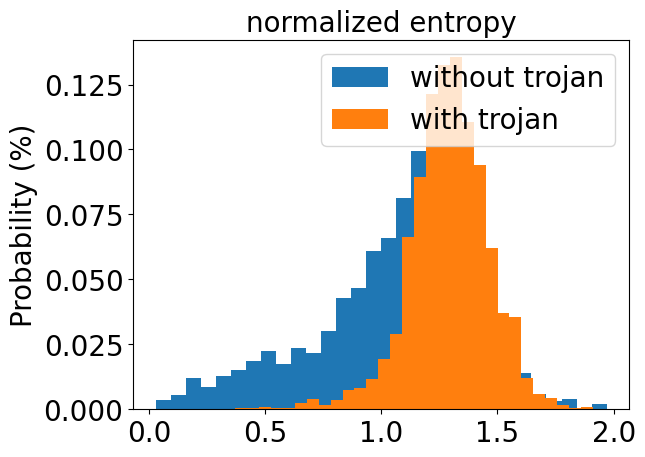}
       \caption{CIFAR10}
    \end{subfigure}%
    \begin{subfigure}[b]{0.5\textwidth}
       \centering 
       \includegraphics[width=\linewidth]{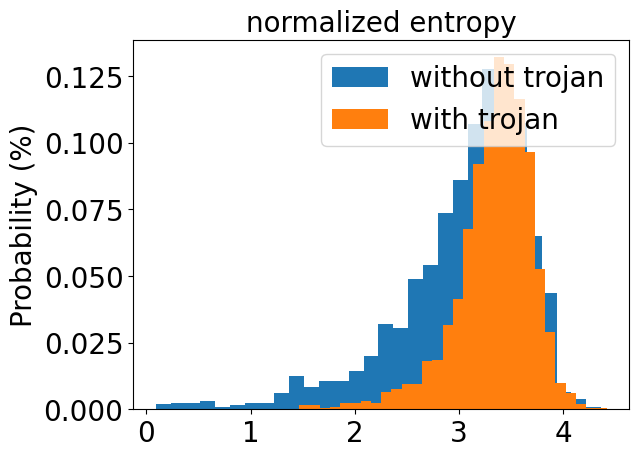}
       \caption{CIFAR100}
    \end{subfigure}
    \caption{Resilience to STRIP}
      \label{fig:example}
  
\end{minipage}%
\begin{minipage}{.5\textwidth}
  
    \centering
    \begin{subfigure}[b]{0.5\textwidth}
       \centering 
       \includegraphics[width=\linewidth]{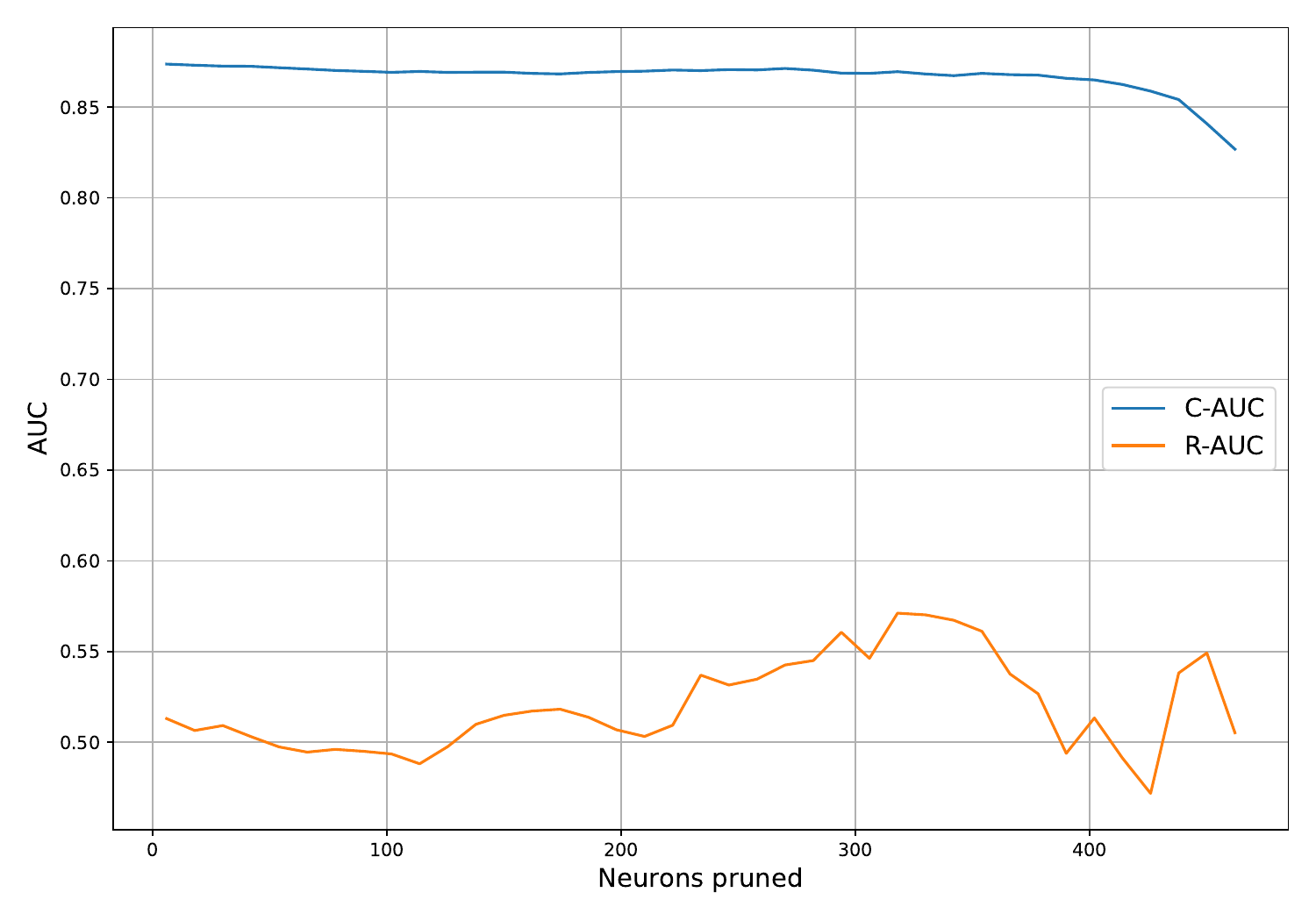}
       \caption{CIFAR10}
       \label{fig:fp_cifar10}
    \end{subfigure}%
    \begin{subfigure}[b]{0.5\textwidth}
       \centering 
       \includegraphics[width=\linewidth]{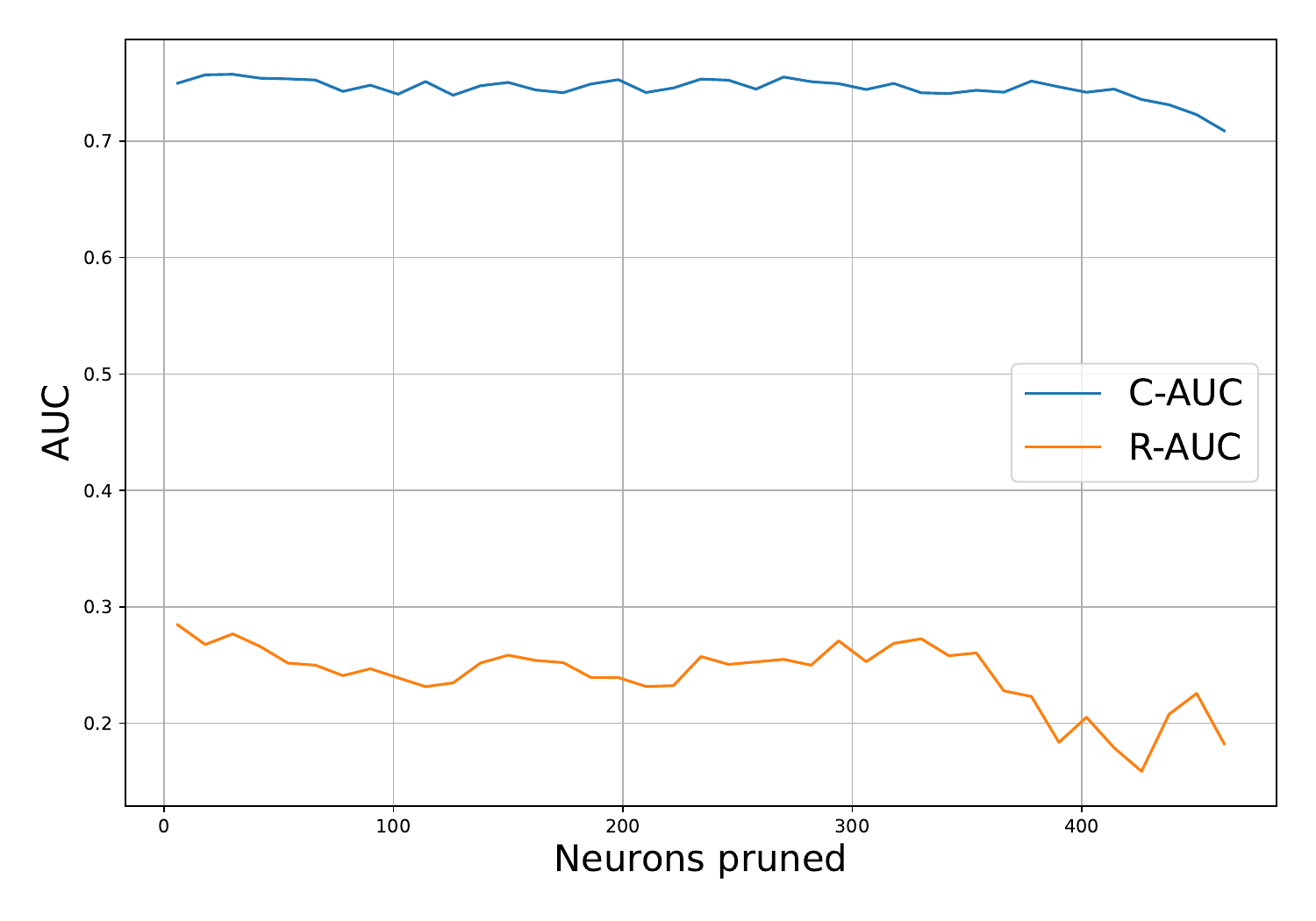}
       \caption{CIFAR100}
       \label{fig:fp_cifar100}
    \end{subfigure}
    \caption{Resilience to Fine-Pruning}
      \label{fig:fp_resilience_figure}
  
\end{minipage}
\end{figure}

\textbf{STRIP} \cite{gao2019strip} serves as a method for detecting threats using a model. It tests the model by adding slight changes to the input image using various safe images from different categories. STRIP signals a warning if the model's predictions don't change much, which means the outcome has low variability. This suggests the model might be compromised. Our attack method mimics the behaviour of unaffected models, showing a similar range of variability in results, as depicted in Figure \ref{fig:example}.

\textbf{Fine-Pruning} \cite{liu2018fine} takes a different approach by examining individual neurons. Focusing on a particular layer, it evaluates how neurons react to a collection of safe images to identify inactive neurons, which are believed to be connected to the hidden threat. These neurons are then carefully removed to reduce the risk. We applied Fine-Pruning to our models and charted the performance of the network, both under normal conditions and during attacks, as we removed neurons. The results can be seen in Figure \ref{fig:fp_resilience_figure}.




  
\begin{figure}[htb]
  \centering
  \begin{subfigure}[b]{0.45\textwidth}
    \includegraphics[width=\linewidth]{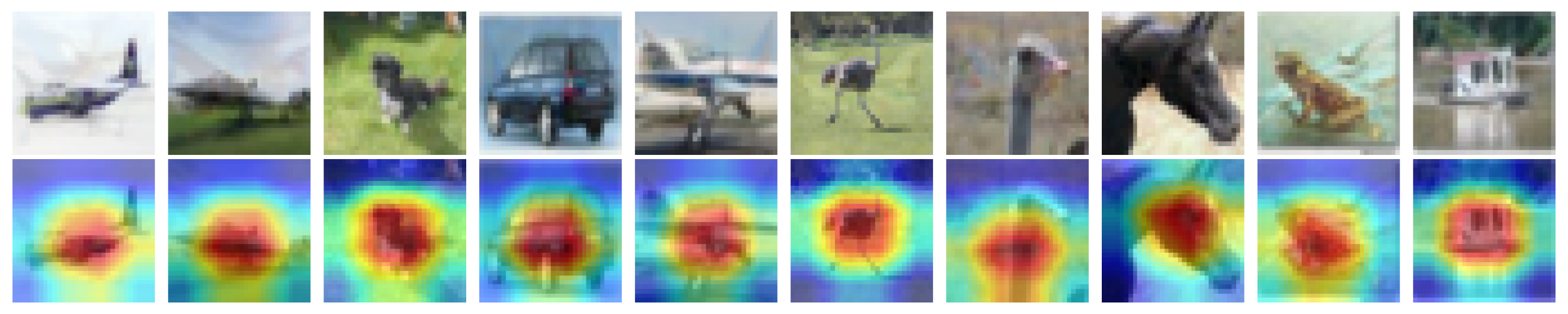}
    \caption{Clean model}
    \label{fig:gradcam_cifar10}
  \end{subfigure}
  \hfill 
  \begin{subfigure}[b]{0.45\textwidth}
    \includegraphics[width=\linewidth]{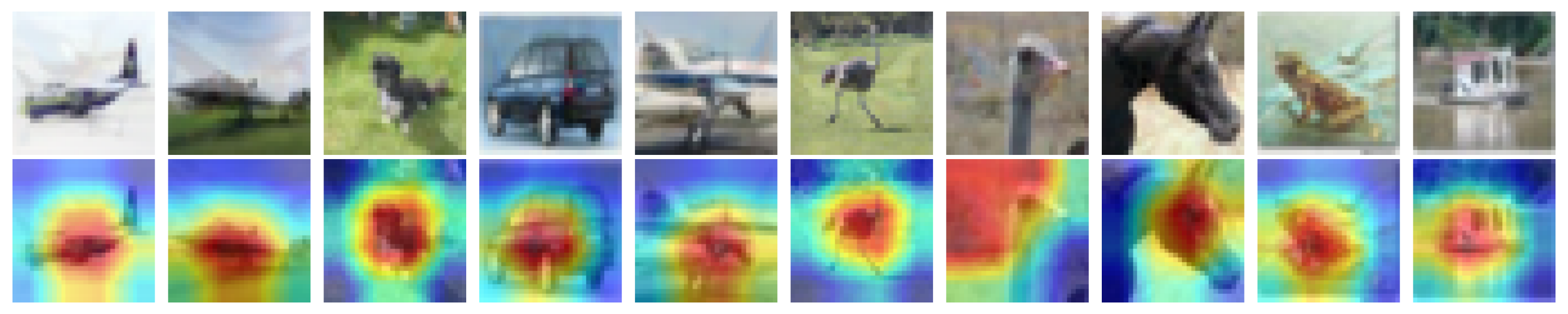}
    \caption{Backdoored model}
    \label{fig:gradcam_cifar100}
  \end{subfigure}
  \caption{Resilience to GradCAM}
  \label{fig:gradcam}
\end{figure}

We also tested the effectiveness of our attack against defenses that use \textbf{GradCAM}\cite{selvaraju2017grad}, a tool that helps understand how models make decisions by generating a heatmap. This heatmap shows which parts of an image are most important for the model's final decision. GradCAM is particularly good at spotting very small, harmful changes in images because these changes usually stand out on the heatmap. However, our method of attack changes the whole image in such a way that GradCAM can't spot these alterations. When we compared the heatmap of a normal model to one from a model affected by our attack, they looked very similar (figure \ref{fig:gradcam}). This similarity suggests that our attack can evade detection by defenses based on GradCAM.

\section{Limitations}
\label{appendix:limitations}
There is a limitation between the number of classes $K$ and the number of samples $|D_a|$. As we want to add $K$ poisoned samples, for any selected samples $x\in X_{j},1 \leq j \leq K$, for any number of classes $K$, we are adding $K^2$ samples to $D^{Out}_{p}$. Moreover, the total size of $D^{Out}_{p}$ should be $\frac{1}{2}$ of available dataset $D_a$. Hence, we have this limitation: $2K^2 \leq |D_a|$.

\renewcommand{\arraystretch}{2}

\begin{table*}[ht]
    \centering
    \caption{Evaluation of the GET attack on the T-Imagenet dataset under various defense mechanisms. The results demonstrate the model's resilience when no defense is applied and its performance across different defense strategies, including NAD, ABL, ANP, SAU, I-BAU, NPD, and RNP. Performance metrics are shown as Benign-AUC/Poison-AUC, where Benign-AUC reflects the accuracy on clean data and Poison-AUC reflects the accuracy under attack conditions. This figure illustrates the impact of class limitations and sample availability as discussed in Section~\ref{appendix:limitations}.}

    \vspace{5pt}
    \setlength{\aboverulesep}{1pt}
    \setlength{\belowrulesep}{1pt}
    \renewcommand{\arraystretch}{1.5} 
    \resizebox{\linewidth}{!}{
        \begin{tabular}{@{}cccccccc@{}} 
        \specialrule{1pt}{0pt}{0pt} 
        \noalign{\smallskip}
         \multirow{2}{*}{\textbf{No Defense}} & \multicolumn{7}{c}{\textbf{Defenses \small\textbf{({Benign-AUC} / Poison-AUC)}}} \\
        \cmidrule(lr){2-8}
        & NAD & ABL & ANP & SAU & I-BAU & NPD & RNP \\        
        \specialrule{0.25pt}{0pt}{0pt} 
         \textbf{39.1 / {7.3 }} & \textbf{43.6 / {4.2 }} & \textbf{42.2 / {17.2}} & \textbf{34.8 / {19.9}} & \textbf{34.5 / {15.1} } & \textbf{37.2 / {18.8}} & \textbf{39.9 / {17.2}} & \textbf{43.5 / {15.7}} \\
        \specialrule{1.5pt}{0pt}{0pt}
        \end{tabular}}
    \label{tab:tiny_imageNet}
\end{table*}

\section{Implementation Detail}
\label{apendix:implementation_detail}
We employ various ResNet variants, ViT-B-16, Wide-ResNet-40-4, and VGG16, categorized either as \textbf{Victim} or \textbf{Surrogate} models. Each model starts with an initial learning rate of 0.01, adjusted using a cosine decay scheduler, with Adam as the chosen optimizer. This selection is predicated on Adam's efficiency in managing sparse gradients. Our setup tests the resilience of these models under a backdoor attack methodology, ensuring rigorous conditions to thoroughly assess the effectiveness of our attacks. The broad applicability and the critical examination of our backdoor attack methodology are underscored, providing a profound insight into the robustness of neural networks against sophisticated threats.

\section{Societal Impact}
\label{appendix:socialImpact}

The societal impact of this research is significant, as it directly relates to safety-critical systems in fields such as autonomous driving and medical diagnostics. By demonstrating the vulnerabilities of machine learning systems to backdoor attacks, the paper sheds light on potential risks that could lead to severe consequences, including threats to human lives. The research underscores the urgent need for robust security measures and defense mechanisms to protect these systems. In the conclusion, the authors acknowledge these risks, contributing to a crucial dialogue on the ethical implications of AI and machine learning research. The discussion aims to foster a deeper understanding of the potential negative impacts, promoting a more responsible approach to the development and deployment of AI technologies in sensitive and impactful areas.

\end{document}